\documentclass{article} % For LaTeX2e
\usepackage[dvipsnames]{xcolor}

\usepackage{iclr2023_conference,times}

\usepackage{amsmath,amsfonts,bm}

\def\eqref#1{equation~\ref{#1}}
\def\1{\bm{1}}

\DeclareMathAlphabet{\mathsfit}{\encodingdefault}{\sfdefault}{m}{sl}
\SetMathAlphabet{\mathsfit}{bold}{\encodingdefault}{\sfdefault}{bx}{n}

\usepackage{hyperref}
\usepackage{url}
\usepackage{todonotes}

\usepackage{amsmath}
\usepackage{amssymb}
\usepackage{mathtools}

\usepackage{hyperref}
\usepackage{cleveref}
\usepackage{booktabs}
\usepackage{styles/annotate_equations}
\usepackage[export]{adjustbox}
\newcommand{\comment}[1]{}

\DeclareMathOperator{\ppl}{ppl}
\newtheorem{phenomena}{Phenomenon}

\title{Scaling Laws for Generative Mixed-Modal Language Models}

\iclrfinalcopy
\author{Armen Aghajanyan\thanks{Equal contribution} \hspace{0.0003em} $^{\dagger}$, \hfill Lili Yu$^{*\dagger}$, \hfill Alexis Conneau$^{\dagger}$, \hfill  Wei-Ning Hsu$^{\dagger}$
\AND Karen Hambardzumyan$^\diamondsuit$, \hfill  Susan Zhang$^{\dagger}$, \hfill Stephen Roller$^{\dagger}$, \hfill Naman Goyal$^{\dagger}$
\AND Omer Levy$^{\dagger}$ \&  Luke Zettlemoyer$^{\dagger, \heartsuit}$ \\ \\
\hspace{10em} FAIR$^{\dagger}$, University of Washington$^\heartsuit$, YerevaNN$^\diamondsuit$ \\
\hspace{14em}    \texttt{armenag@meta.com}}

\begin{document}

\comment{
    P0: Correct Log PPL in empirical data
    P0: Add goodness of fit plots somewhere...
    P0: Double check experiments in optimal batch size, don't trust jupyter notebook, try to recover AWS logs.
    P0: Lost MM Image experiment, due to azcopy command... Try to retrain before arxiv?
    P1: Should we discuss, fickleness of L-BFGS? Chinchilla doesn't mention anything
    P1: Do we need real definition of modality? I think it's obvious we can remove.
    P1: Dont like title
}
\maketitle

\begin{abstract}
    Generative language models define distributions over sequences of tokens that can represent essentially any combination of data modalities (e.g., any permutation of image tokens from VQ-VAEs, speech tokens from HuBERT,  BPE tokens for language or code, and so on). 
    To better understand the scaling properties of such mixed-modal models, we conducted over 250 experiments using seven different modalities and model sizes ranging from 8 million to 30 billion, trained on 5-100 billion tokens. We report new mixed-modal scaling laws that unify the contributions of individual modalities and the interactions between them. Specifically, we explicitly model the optimal synergy and competition due to data and model size as an additive term to previous uni-modal scaling laws. We also find four empirical phenomena observed during the training, such as emergent coordinate-ascent style training that naturally alternates between modalities, guidelines for selecting critical hyper-parameters, and connections between mixed-modal competition and training stability. Finally, we test our scaling law by training a 30B speech-text model, which significantly outperforms the corresponding unimodal models. Overall, our research provides valuable insights into the design and training of mixed-modal generative models, an important new class of unified models that have unique distributional properties. 
%which have unique distributional properties and are increasingly important in the development of unified models.
\end{abstract}

\section{Introduction}
Generative language models have been developed for a wide range of data modalities, including natural language text \cite{gpt3}, code \citep{codex, INCODER}, images \citep{DALLE, RA_CM3}, and molecules or proteins \citep{BARTSMILES, hsu2022learning}. Recent work has also introduced unified models \citep{CM3,gato,ofa,Merlotreserved} that can simultaneously model multiple modalities. One advantage of generative modeling in these cases is that the models scale well in practice; adding data, compute, or parameters typically improves model quality. These scaling trends have been carefully studied for uni-modal models~\citep{kaplan2020scaling, Chinchilla} and some recent work focuses on pairs of modalities~\citep{speech_alexa_scaling_laws, image_text_scaling}. However, the scaling behavior of larger number of modalities remains largely unstudied.

We present an extensive empirical study of scaling laws for mixed-modal generative language models over tokens. We assume that every modality can be represented as a sequence of tokens (e.g. VQ-VAEs for images~\citep{taming} or HuBERT for speech~\citep{hubert}). With this assumption, we can train a single discrete language model to represent data with arbitrary subsets of modalities presented in arbitrary orders. Such mixed-modal models are very general, but it is an open question the extent to which scale alone will be enough to overcome the inherent competition that comes as we add more modalities to a single model. 

Through extensive experimentation, including over 250 individual experiments with seven modalities and model sizes ranging from 8 million to 30 billion, we have identified a scaling law that reflects the contributions of individual modalities and an additional term that captures the interaction between modalities (whether it be one of competition or synergy). We develop mixed-modal scaling laws that directly model competition between modalities and correctly predict data and model regimes where competition between modalities during training progresses into synergy. Specifically, we showed that our scaling laws correctly predicted the compute regime (30B model size, 45B token size), where we saw the complete reduction of modality competition for the Speech and Text modalities.

We also report a number of new empirical phenomena that arise during the training of mixed-modal models, including the tendency for the models to prioritize the optimization of a single modality at different stages of training. Our findings demonstrate that these phenomena can be primarily explained through the scaling law of interaction within the mixed-modal model. Additionally, we present new insights and guidelines for how to set key hyperparameters based on the terms of our scaling laws when optimal uni-modal hyper-parameters are known.

% Through devising a unified tokenization strategy for fixed set of multiple modalities and 

Our contributions are the following:
\begin{itemize}
    \item We develop neural scaling laws for mixed-modalities models that include text, speech, images, code, and their numerous couplings.
    \item We discover a set of scaling laws describing the competition between arbitrary modalities.
    \item We provide a simple recipe for selecting hyper-parameters in a multi-modal setting when optimal uni-modal hyper-parameters are known.
    \item We uncover correlations between the scaling laws parameters we propose and various training phenomena, including training stability, optimal batch size, and coordinate ascent-like behavior in the optimization process across different modalities.
\end{itemize}

\section{Related Work}
Neural scaling laws quantify the relationship between model size, dataset size, compute budget, and performance, when training neural networks. This concept was introduced by \citet{deep_learning_scaling_2017}, who observed a power law relationship and later scaled to much larger models by  \citet{kaplan2020scaling}. 

\citet{Chinchilla} developed a unified formula for scaling laws, and provided recipes for compute-optimal training by adding data-dependent scaling terms unlike previous power law parameterizations.  
Other researchers have applied these principles to specific tasks and different parameterization of Transformers. \citet{moe_scaling_laws} examined the application of neural scaling laws to Mixture of Experts (MoE) models. \citet{int8, 4_bit_scaling} studied the relationship between scaling laws and lower precision, which refers to using lower-precision data types, such as 16-bit floating point numbers, in neural networks. \citet{scaling_laws_nmt} and \citet{scaling_laws_nmt_ghorbani} applied these principles to Neural Machine Translation (NMT).

Additionally, \citet{image_text_scaling} and \citet{speech_alexa_scaling_laws} examined the application of neural scaling laws to generative language models in different modalities, including image generation and acoustic models. \citet{CLIP_scaling_law} also examined multi-modal training but did not specifically focus on generative models.
To our knowledge, we are the first to investigate the phenomenon of interactions, competition, and interference between multiple modalities during training and provide a recipe for optimal mixed-modal training. 
%This is an important area of research as it allows practitioners to effectively train neural networks on multiple modalities, potentially leading to improved performance on tasks such as image and language generation.

Interestingly, similar competition and scaling phenomenon have been observed for multi-lingual models. \citet{conneau2019unsupervised} observed a ``curse of multilinguality,'' where training in multiple languages can lead to interference between languages, resulting in decreased performance. \citet{goyal2021larger} and \citet{interference_translation} demonstrated that this interference could occur even on models much smaller than the available training data, but scaling up the model size can improve synergy and alleviate interference. These findings align with our findings in the mixed-modal scenario, suggesting that similar principles apply when training on multiple modalities.

\section{Definitions}
\subsection{What is a Modality?}
Modalities are traditionally distinguished by the data source, domain, or sensor affinity. For example, the code domain is typically seen as distinct from text due to the different data involved (e.g., GitHub vs. CommonCrawl). This also applies to auditory or visual modalities, which are captured with different sensors. Yet the decisions are not always clear, for example different languages are often all within the domain of the text. Given that we are studying neural scaling laws across modalities, we aim to have an empirically testable modality definition.

We define $\sigma-$membership of a set of samples, $D_{\alpha}$ through the following membership function.
\begin{equation}
    D_{i} \in D_{j} \iff \mathbb{E}_{x\sim D_i}\left[\mathcal{L}_{D_j}(x)\right] \le \sigma^2 \mathbb{E}_{x\sim D_j}\left[\mathcal{L}_{D_j}(x)\right]
\end{equation}

We empirically define \textbf{modality} by comparing the perplexity of one data set to another. Suppose the perplexity of the secondary data set over the probability distribution of the primary set is greater than $\sigma$ times the mean perplexity of the primary set. In that case, we consider them to be distinct modalities. This definition distinguishes modalities by source, domain, sensor affinity, and language. We use the standard definition of perplexity ($\ppl$). Using this definition with $\sigma=3$, we decided to select seven modalities that we describe in detail below: \texttt{Text}, \texttt{Image}, \texttt{Image-Text}, \texttt{Speech}, \texttt{Speech-Text}, \texttt{Code}, \texttt{Molecules}.

Additionally, we define \textbf{source modality} as the type of token the sample contains, which within our setting will be; \texttt{Text}, \texttt{Speech}, or \texttt{Image}. 

\subsection{Uni-Modal Scaling Laws}
We selected the \citet{Chinchilla} parameterization of scaling laws due to its precise representation of data factors and its additive nature, which allows for easy extension to multiple modalities. This parameterization (Equation~\ref{eq:chinchilla_scaling_law}) describes the loss based on the number of model parameters ($N$) and the number of tokens ($|D|$) through three constituent parts: the minimal achievable loss ($E$), the functional approximation error ($\frac{A_j}{N^{\alpha_j}}$), and the optimization or convergence error ($\frac{B_j}{|D_j|^{\beta_j}}$). These three factors are captured through seven learned parameters, providing a precise description of the loss.

It is well established that the upper bounds for $\beta$ and $\alpha$ are both $\frac{1}{2}$, which provides a clear understanding of how well transformers coupled with gradient descent algorithms scale in relation to the optimal scaling for each modality \citep{Chinchilla}.

\begin{figure}[h]
\vspace{1em}
\begin{equation}
\label{eq:chinchilla_scaling_law}
\mathcal{L}
    \left(\eqnmarkbox[NavyBlue]{N}{N}, 
    \eqnmarkbox[OliveGreen]{D}{D}_{\eqnmarkbox[WildStrawberry]{j1}{j}}\right) 
    = \eqnmarkbox[Plum]{E}{E}_{\eqnmarkbox[WildStrawberry]{j2}{j}} + 
    \eqnmarkbox[Emerald]{func}{\frac{A_j}{N^{\alpha_j}}} + 
    \eqnmarkbox[BurntOrange]{conv}{\frac{B_j}{|D_j|^{\beta_j}}}    
\end{equation}
\annotate[yshift=1em]{above,left}{N}{Number of Model Parameters}
\annotate[yshift=-1em]{below,left}{D}{Dataset}
\annotatetwo[yshift=1em]{above}{j1}{j2}{For Modality $j$}
\annotate[yshift=-2.5em]{below,left}{E}{Minimal Achievable Loss}
\annotate[yshift=1em]{above,right}{func}{Functional Approximation Error}
\annotate[yshift=-0.5em]{below,right}{conv}{Convergence Error}

\vspace{1em}
\end{figure}

\section{Empirical Setting}
    \subsection{Datasets}
        \paragraph{\textbf{Text}} For our text corpus, we use the same data as was used in OPT \citet{OPT} for a total of 180B tokens. This dataset is primarily in English, although it contains other languages, as no explicit language filtering was done.
        \paragraph{\textbf{Image}} For all images, we convert them to discrete tokens using the Make-A-Scene visual tokenizer \citep{make_a_scene}, which gives 1024 tokens from an 8192 vocabulary per image. We select a custom subset of 600 million images across \citet{laion}, and a custom image-text dataset scraped from Common Crawl. We remove all NSFW images and images that contain watermarks. Our \texttt{Image} dataset only contains the image and not the caption for a total of 614 billion tokens.
        \paragraph{\textbf{Image-Text}} We utilize the \texttt{Image} dataset described above but align it with captions available from the image for a total of 690 Billion tokens. We call this our \texttt{Image-Text} dataset.
        \paragraph{\textbf{Speech}} We used a combination of custom web-mined speech data and unlabeled speech in several public datasets. The web-mined speech dataset contains only unlabeled data in the form of long podcasts or news. We follow a series of preprocessing steps to improve the data quality and remove music and sensitive speech data. We also use a LangID model to select English-only speech. Our public data collection covers various speech styles and content topics, including LibriSpeech (Read-Books), CommonVoice in Read-Wiki, VoxPopuli from the Parliament domain, and Spotify Podcast and People's Speech as web speech. Thanks to this combination, our \texttt{Speech} dataset offers a rich diversity. 
        \paragraph{\textbf{Speech-Text}} Many public datasets also come with text aligned with speech. We take ASR and TTS data from Multilingual LibraSpeech and VoxPopuli and form the \texttt{Speech-Text} dataset. 
        \paragraph{\textbf{Code}} We use the InCoder data \citep{INCODER}.
        \paragraph{\textbf{Molecules}} We utilize the Simplified Molecular Input Line Entry System (SMILES, where the chemical’s structure is serialized into a string of symbols) representation from the Zinc dataset prepared by \citet{BARTSMILES}.
    \subsection{Tokenization}

       Our mixed-modal generative models use a unified tokenization over all the modalities mentioned. This tokenizer processes data from all modalities into discrete tokens, which can be processed jointly by our model and trained with a single loss. %This requires us to develop a pretokenization process for data that naturally reside in continuous space, such as image and speech. 
        
        We use a Vector Quantized Variational autoencoders (\textsc{VQGAN} \cite{taming}) model to tokenize image data into discrete tokens. The \textsc{VQGAN} model compresses each image into a grid of image tokens, where an encoder encodes each token into a vector.  This process reduces the context size of the transformer by a factor of $3*X^2$, where $X$is the spatial reduction rate, or patch size, and 3 is the number of image channels. Online clustering is then performed, mapping each vector to the nearest entry of a learned codebook. We use a variant of the \textsc{VQGAN} from \citet{make_a_scene}, which has a spatial reduction of 8 and a codebook size of 8192. This model is trained with extra perceptual losses to speciﬁc image regions, such as faces and salient objects, which improves the ﬁdelity of the generated images. To be most effective in the language model stage, the visual tokenizer needs to effectively represent a image, and the correlated decoder needs to reconstruct the generated image tokens into high quality image data. We benchmark various image pretokenizers for those properties in Appendix \ref{sec:image_tokenization}. 
        
        We use a Hidden-Unit BERT (HuBERT) \cite{hubert} model for tokenizing our speech data. HuBERT is a self-supervised learning (SSL) model. It is trained to predict a masked subset of the speech signal using a mask language model objective, and has been found to be effective in learning a combined acoustic and language model over the continuous speech inputs. An offline clustering step to then used to generate discrete units. We use the \textsc{Base} HuBERT model in our work (model and training details see the appendix \ref{sec:speech_tokenization}). 
        % This model comprises a convolutional encoder and 12 layer Transformer, each with an embedding dimension of 768, a feed-forward layer dimension of 3072, and 12 self-attention heads. 
        % Pre-training of the model has been performed on 32 GPUs over three iterations, with 400K updates per iteration. 
        % The training data consists of 221K hours of unlabeled speech from multilingual Librispeech~\cite{pratap2020mls}, Common Voice~\cite{ardila2019common}, and VoxPopuli~\cite{wang2021voxpopuli} in eight languages (English, Spanish, French, German, Dutch, Italian, Polish, Portuguese). The MFCC/6-th layer feature from iteration 1 and the 9-th layer feature from iteration 2 are used as targets, with codebook sizes of 100/500/1000, respectively, following the methodology outlined in \cite{lee2021direct}. 
        The final HuBERT units are generated through K-means clustering of the third iteration feature at the last layer, with a codebook size of 2000. Our HuBERT model encodes audio at 50Hz, and we compress a 16kHz audio by about 120 times, while effectively retaining specch information (Analysis see \ref{sec:speech_tokenization}).
        % A typical 16kHz audio with a bit depth of 16 has a bitrate of 64kbps. HuBERT encodes audio at 50Hz with a codebook size of 2000, resulting in a bitrate of 548bps. The effective compression rate is roughly 117x.

        Finally, we randomly sample 10 million sentences from all the data sets mentioned above and train a BPE model, where image and speech tokens take up a single token. We do an additional digit splitting for a vocab size of $2^{16}$ \citep{BPE}.
        
    \subsection{Model Architecture}
        We study the family of decoder-only models described in GPT-3 \citet{gpt3} and OPT \citet{OPT}. We limit ourselves to training up to 6.7 billion-parameter models for all our uni-modal and bi-modal scaling laws and train up to 30B parameters to measure the generalizability of our scaling laws. For completeness, we present model architecture and their respective sizes in Table~\ref{table:opt_model_desc}.
        We use learned positional encodings across all model architectures.

% We report the number of layers (#L), the embedding size ,  number of attention heads (#H), and the dimension of each attention head()
    \subsection{Causal Masking Objective}
        Instead of the traditional left-to-right causal language modeling objective, we use the causal masked objective from \citet{CM3}. This provides a form of bidirectional context for  sequence infilling, and also supports more aggressive generalization. For example, causally masked models trained only on data with text followed by images can still flip the ordering to generate images from text, since they were not strictly trained to predict tokens left to right.
%        The main benefit of using the causal masked objective is that it allows us to model conditional probabilities of modalities in both ways. Modeling $p(\texttt{text}, \texttt{image})$ with the standard causal language modeling loss gives us the ability to model $p(\texttt{image} | \texttt{text})$ but not $p(\texttt{text} | \texttt{image} )$. The causal masking alleviates this by allowing us to recover $p(\texttt{text} | \texttt{image})$ by properly masking out the input sequence, therefore allowing us to significantly reduce the number of training runs we need to discover proper multi-modal scaling laws.
        Recent work also shows that this masking does not hurt language modeling performance or the generative capacity of the models \citep{INCODER, bavarian2022efficient}. We provide additional support for this claim in \S~\ref{sec:cm_vs_c}.
    \subsection{Training Procedure}
        \label{sec:training_proc}
        All models were trained using the metaseq\footnote{https://github.com/facebookresearch/metaseq} code base, which includes an implementation of causal masking \cite{OPT}. The training used the PyTorch framework \cite{pytorch}, with fairscale to improve memory efficiency through fully sharded model and optimizer states \cite{fairscale}. The training also uses Megatron-LM Tensor Parallelism \citet{megatron} to support large model runs, and we use bf16 \cite{bf16} to improve training stability. Given the large volume of data, we performed a single epoch of training, using each training document once. The batch size per GPU was determined based on the total world size of the experiment, the level of model parallelism, and the total target batch size in terms of the number of tokens. To ensure stable training, we applied gradient clipping with a maximum norm of 1.0 and used the Adam optimizer with $\beta_1 = 0.9$, $\beta_2 = 0.98$ \cite{ADAM}. We used the built-in polynomial decay learning rate scheduler in MetaSeq with 500 warmup updates and the end learning rate set to 10\% of the peak learning rate.

        We tracked all experiments using the Aim experiment tracker \citep{AIM}. To ensure consistent training strategies across our experiments, we implemented a model restart policy using the Aim experiment tracker and callbacks. Specifically, if training perplexities do not decrease after 500 million tokens, the training run is restarted with a reduced learning rate with a factor of 0.8 of the current time step. This policy helps remove variance in the scaling laws due to differences in training procedures and allows us to scale up the number of asynchronous experiments significantly.

        All experiments were conducted in a two-month time frame with a cluster of 768 80GB A100 GPUs. The majority of experiments used 64 GPUs at a time.
\section{Scaling Laws}
    \subsection{Uni-Modal Scaling Laws}
    We first aim to discover scaling laws for each of the individual modalities we listed above. We train seven different model sizes, from 8 million to 6.7 billion, on seven different modalities on three different dataset sizes (5B, 10B, 100B). 

\begin{figure}[h]
        \centering
        \includegraphics[width=\textwidth]{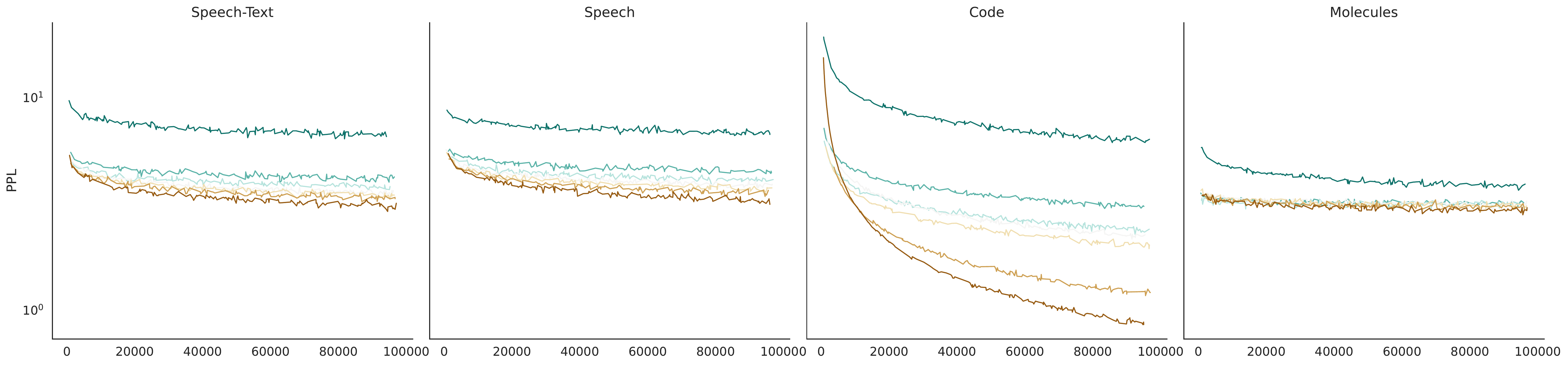}
        \includegraphics[width=\textwidth]{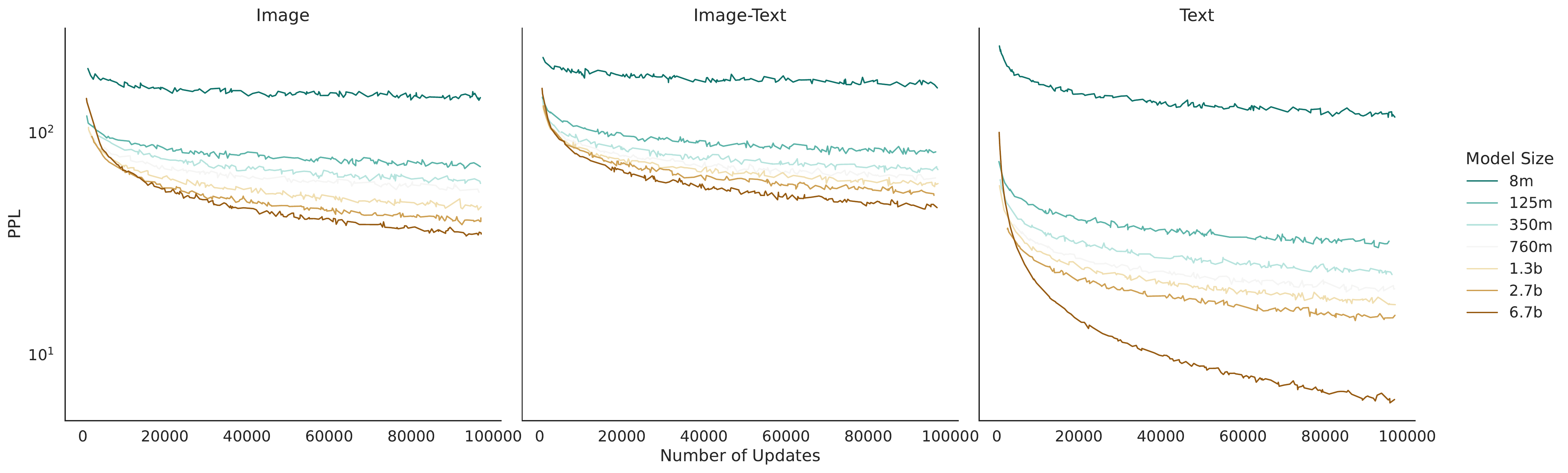}
        \label{fig:monomodal_scaling}\caption{Single modality training curves for 100B tokens across a wide range of model sizes. Different modalities exhibit wildly different training dynamics.}
    \end{figure}

\begin{figure}[h]
        \centering
        \includegraphics[width=\textwidth]{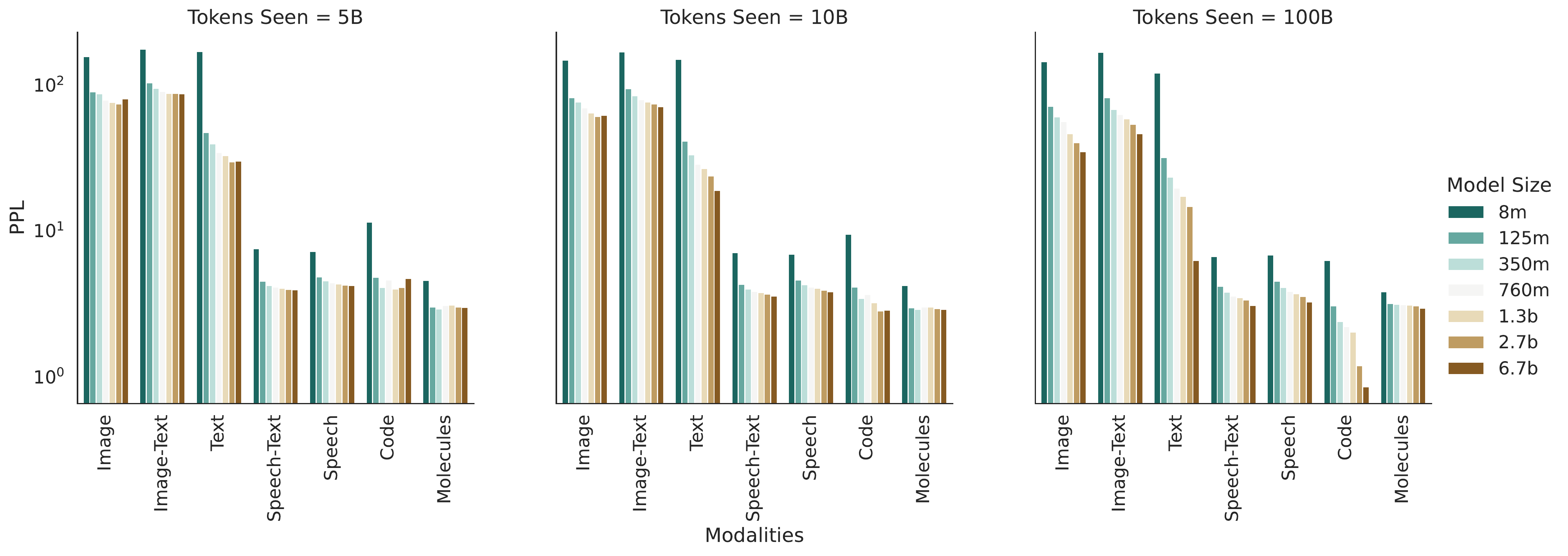}
        \caption{Empirical scaling properties across both data and model size scale for the uni-modal setting.}
        \label{fig:moomodal_scaling_by_D}
    \end{figure}

    In Figure~\ref{fig:monomodal_scaling}, we share the training curves for all modalities and model sizes for the largest data size (100B tokens), and the final performance of all models in Figure~\ref{fig:moomodal_scaling_by_D}. Overall, we see that scaling dynamics are fundamentally different across modalities, scale, and dataset size (which further reinforces our selection of dataset-size-dependent parameterization of scaling laws). 
    
    For each modality, we fit the seven parameters from Equation~\ref{eq:chinchilla_scaling_law}, following the procedure in \citet{Chinchilla}. Specifically, we minimize 
    \begin{equation}
        \label{eq:minimizing_scaling_law}
        \min_{a_j,b_j,e_j,\alpha_j,\beta_j} = \sum_{\texttt{run i in modality j}} {Huber}_{\sigma=0.03}\left[LSE(a_j - \alpha_j \log{N_i}, b - \beta \log{D_i}, e_j) - L_i\right]
    \end{equation}
    We then set $A_j=e^{a_j}$, $B_j=e^{b_j}$, $E_j=e^{e_j}$. In order to identify the optimal minima, we followed the method outlined by \citet{Chinchilla} and employed the L-BGFS algorithm on the same grid of initialization values. Our only deviation was using a higher value for the Huber loss parameter $\sigma$, which was necessary for generalization to held-out data in our multi-modal setting. The optimal values obtained were not located on the boundaries of the initialization grid.
    
    \begin{table}[h]
        \centering\small
        \begin{tabular}{lrrrrrrr}
        \toprule
        {} &    Code &  Image-Text &  Image &  Molecules &  Speech-Text &  Speech &    Text \\
        \midrule
        A     &  611.91 &      320.51 & 340.96 &     158.19 &       180.68 &  154.45 &  492.51 \\
        B     & 4484.08 &      658.31 & 875.30 &     189.36 &       234.13 &  205.10 & 1987.40 \\
        E     &    0.16 &        2.47 &   2.84 &       2.39 &         2.69 &    3.02 &    2.42 \\
        $\alpha$ &    0.37 &        0.12 &   0.13 &       0.37 &         0.32 &    0.31 &    0.18 \\
        $\beta$  &    0.32 &        0.11 &   0.13 &       0.26 &         0.24 &    0.24 &    0.22 \\
        \bottomrule
        \end{tabular}
        \caption{Uni-Modal scaling law parameters fit to Equation~\ref{eq:chinchilla_scaling_law} (Chinchilla Scaling Law).}
        \label{tab:monomodal_scaling_laws}
    \end{table}

    The scaling laws for each modality are presented in Table~\ref{tab:monomodal_scaling_laws}. The parameters for each modality vary significantly. Some modalities, such as \texttt{Code} and \texttt{Molecules}, demonstrate more efficient use of the power of scale compared to others, such as \texttt{Image}. Our coefficients for \texttt{Text} are similar to those reported by Chinchilla, although it should be noted that we used a different dataset for our analysis. This accounts for any differences in the results.
\subsection{Bi-Modal Scaling Laws}
    \label{sec:bimodal_scaling}
    We also estimate scaling laws for training on two modalities: $\mathcal{L}\left(N, D_i, D_j\right)$, where $N$ represents the model size, and $D_i$ and $D_j$ represent the two datasets being used. In the case where $D_i$ and $D_j$ are completely independent and have no mutual information between them, we expect the minimal achievable loss to be the average of the two monomodal scaling laws, given by $0.5 * \left[\mathcal{L}\left(\infty, D_i\right) + \mathcal{L}\left(\infty, D_j\right)\right]$. This is because we are averaging over the loss and subsampling both test datasets equally ($|D_i| = |D_j|$). On the other hand, if there is some form of mutual information present between $D_i$ and $D_j$, we can expect the loss to be reduced by some maximal factor $C_{i,j}$. When considering finite model size and data regimes, there will be competition between the function approximation and optimization processes, which can be modeled using the same form as in Equation~\ref{eq:chinchilla_scaling_law}. We present our scaling law for mixed modal models in Equation~\ref{eq:mm_scaling_law}.
    
    \begin{figure}[h]
        \vspace{1.5em}
        \begin{equation}
        \label{eq:mm_scaling_law}
        \mathcal{L}\left(N, D_i, D_j\right) 
            = \left[\eqnmarkbox[Plum]{additive}{
                \frac{\mathcal{L}\left(N, D_i\right) + \mathcal{L}\left(N, D_j\right)}{2}}\right]
                - \eqnmarkbox[NavyBlue]{Cij}{\mathcal{C}_{i,j}} + 
                \eqnmarkbox[Emerald]{func1}{\frac{A_{i,j}}{N^{\alpha_{i,j}}}} + 
                \eqnmarkbox[BurntOrange]{conv1}{\frac{B_{i,j}}{|D_{i}| + |D_{j}|^{\beta_{i,j}}}} 
        \end{equation}
        \annotate[yshift=1em]{above,right}{Cij}{Maximum Level of Synergy}
        \annotate[yshift=-0.5em]{below,left}{func1}{Competition in Functional Approximation}
        \annotate[yshift=-1.5em]{below,left}{conv1}{Competition in Optimization Process}
        \annotate[yshift=0.5em]{above,left}{additive}{Loss if Datasets Were Modeled Independently}
        \vspace{1.5em}
    \end{figure}

    An additional benefit to this parameterization is the additive or linear nature, which allows us to extend our parameterization to $n$-modal scaling laws. 
    
    \subsubsection{Experimental Results}
    We selected seven different pairs: \texttt{Image-Text|Code}, \texttt{Image-Text|Speech-Text}, \texttt{Image-Text|Text}, \texttt{Speech|Text}, \texttt{Code|Text}, \texttt{Molecule|Code}, and \texttt{Speech|Code}. While other couplings are available, we cannot do an exhaustive sweep due to computational constraints. We selected these pairs to maximize variety. For example, while \texttt{Code|Text} is known to perform well, \texttt{Image-Text|Code} may not offer as much benefit.

    We create a dataset for each coupling and token target count where each subdataset contributes 50\% of the tokens. We train using the same hyper-parameters as the uni-modal trainings and fit the scaling laws per modality coupling using the same procedure and optimization process (Equation~\ref{eq:minimizing_scaling_law}).

\begin{figure}[h]
        \centering
        \includegraphics[width=\textwidth]{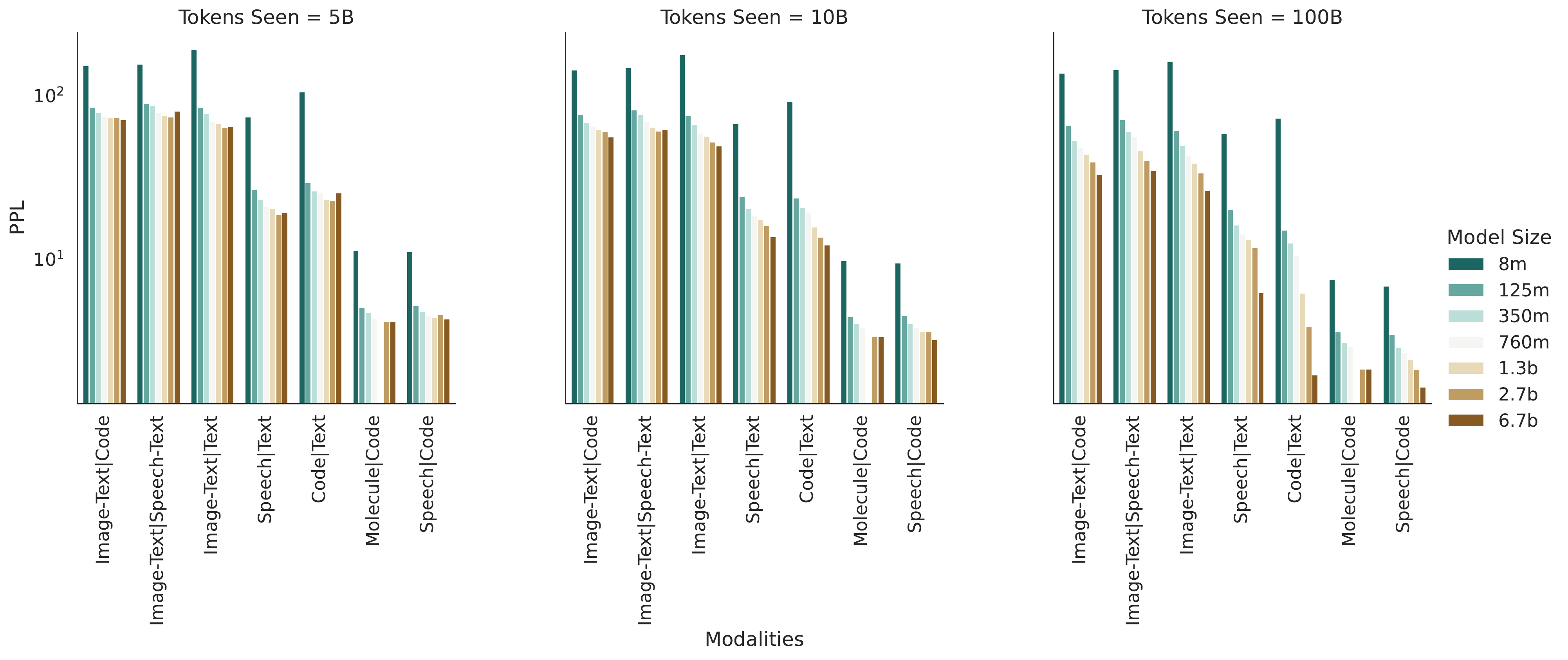}
        \caption{Empirical scaling properties across both data and model size scale for the multi-modal setting.}
        \label{fig:multimodal_scaling_by_D}
    \end{figure}

    We present the empirical results in Figure~\ref{fig:multimodal_scaling_by_D}.
    
    \subsubsection{Breaking The Competition Barrier}
\begin{figure}
        \begin{minipage}[b]{1.0\linewidth}
            \centering
            \includegraphics[width=0.49\linewidth]{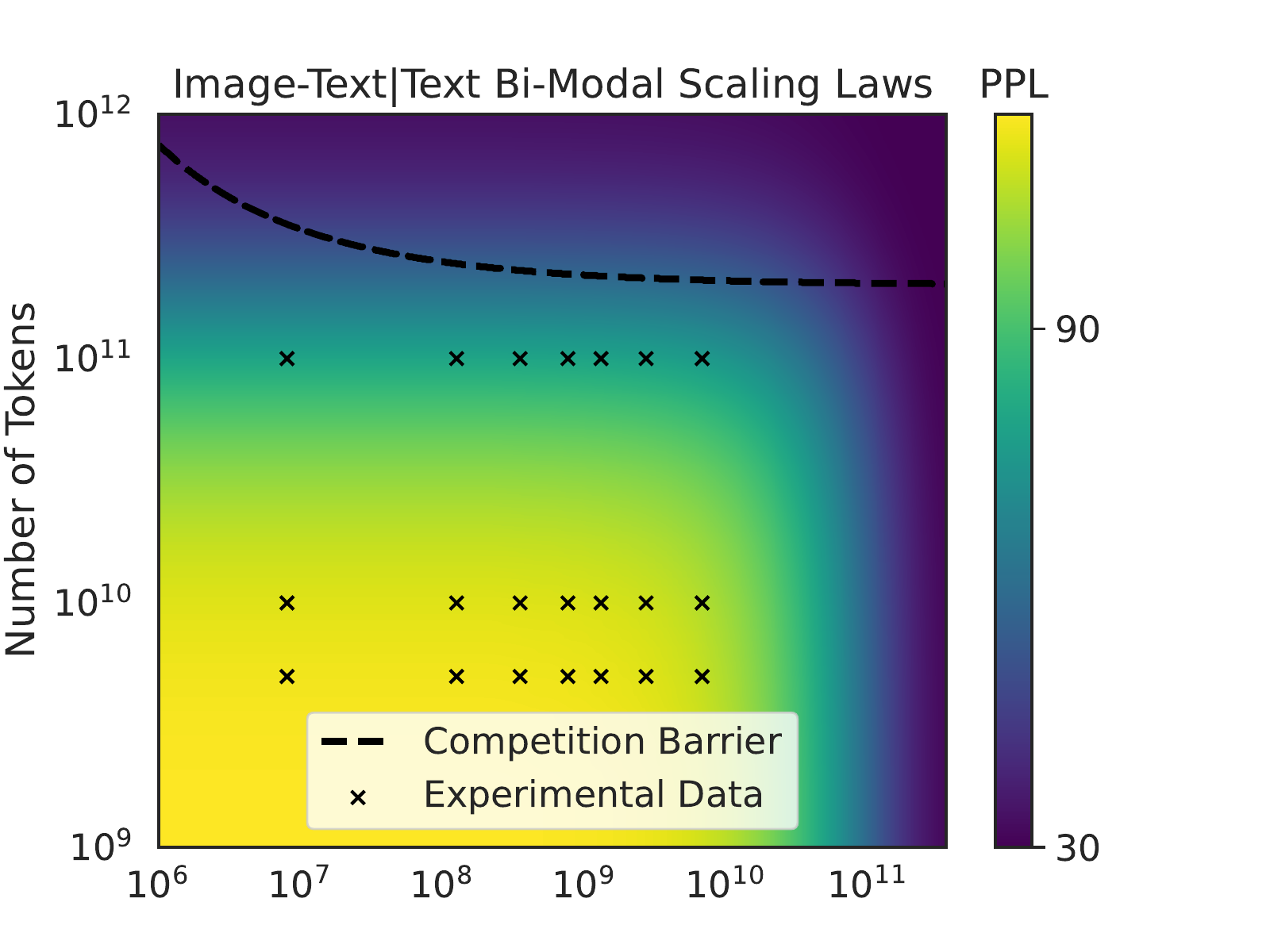}
            \includegraphics[width=0.49\linewidth]{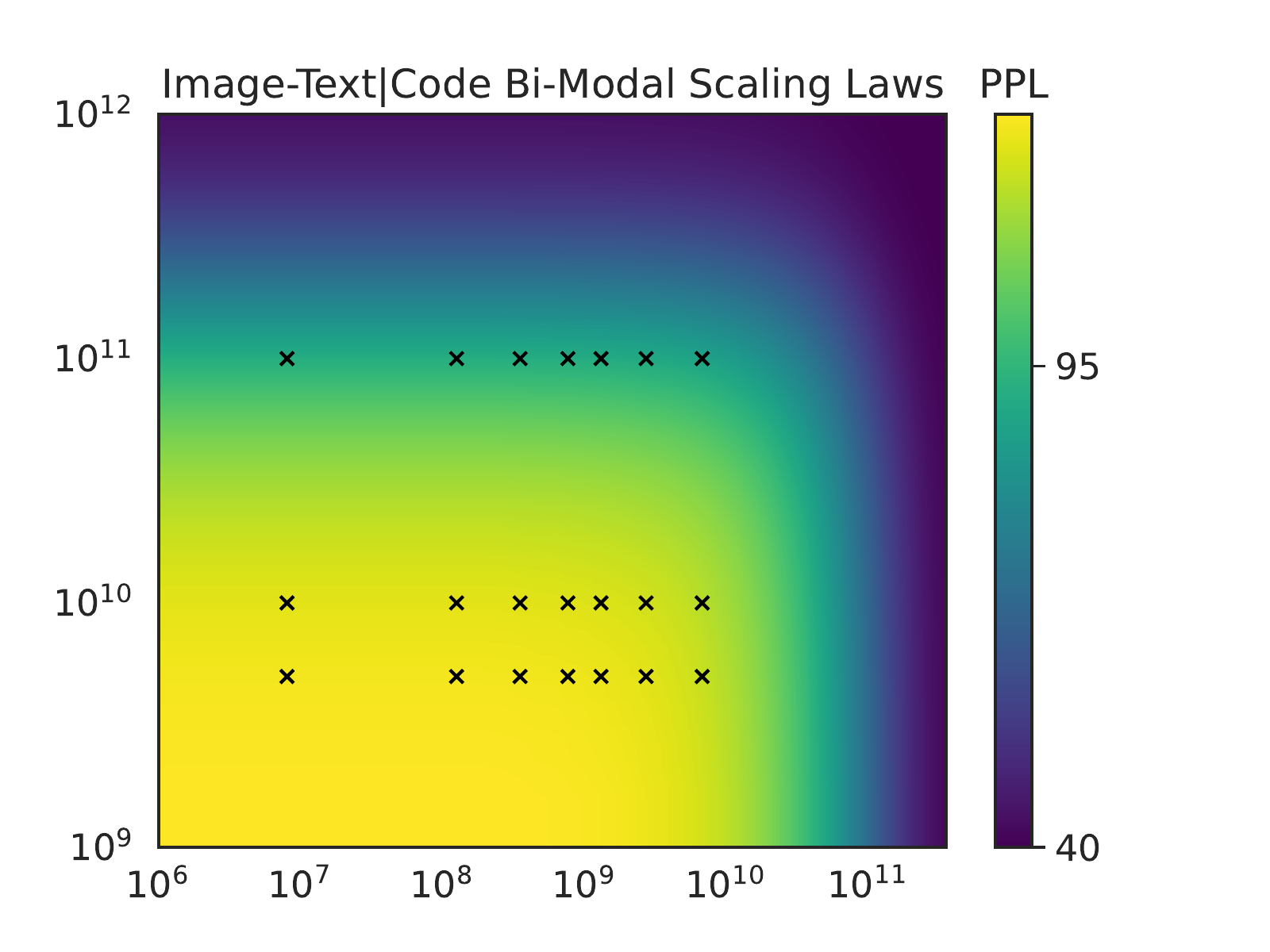}
        \end{minipage}
        \begin{minipage}[b]{1.0\linewidth}
            \centering
            \includegraphics[width=0.49\linewidth]{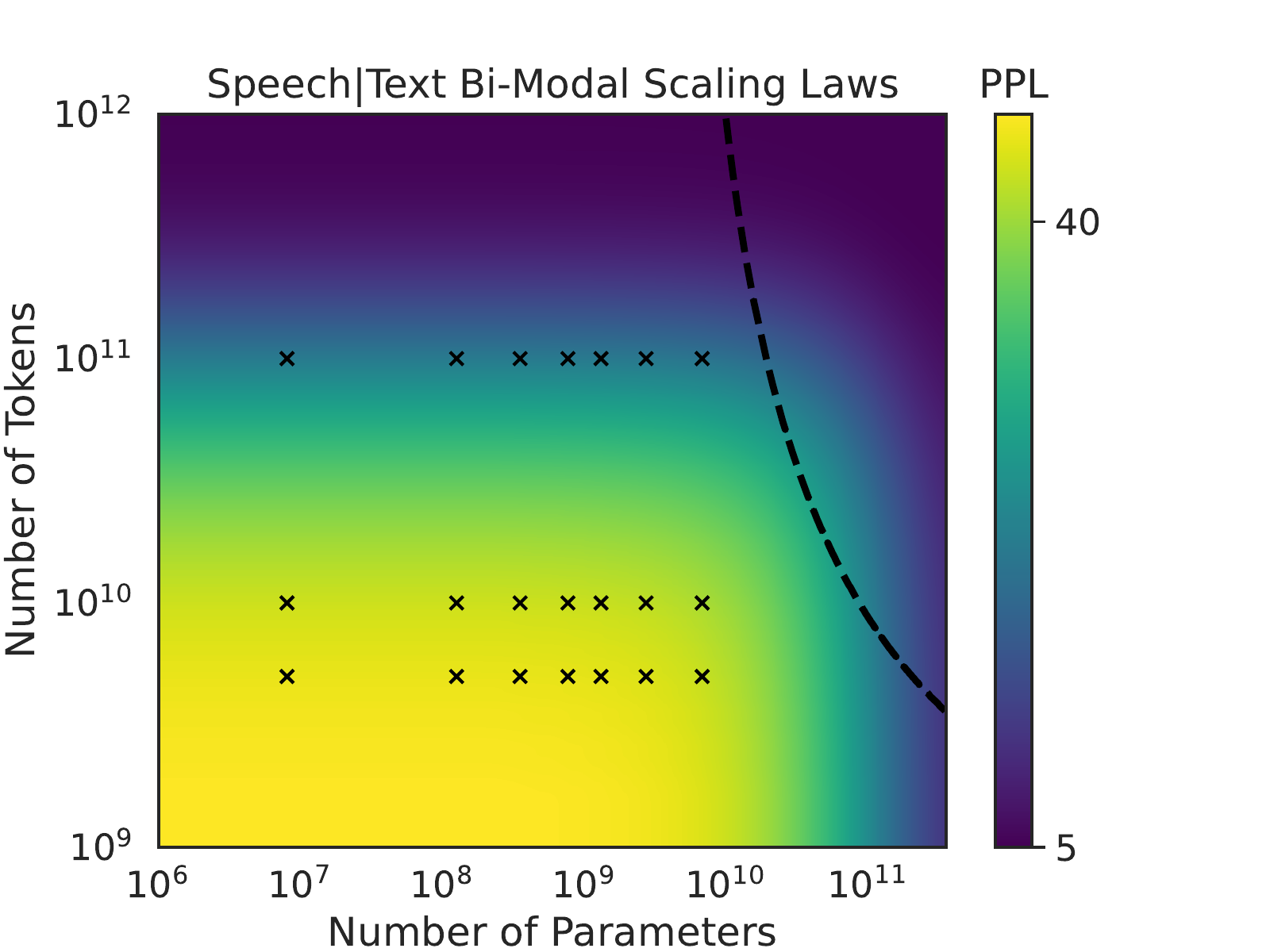}
            \includegraphics[width=0.49\linewidth]{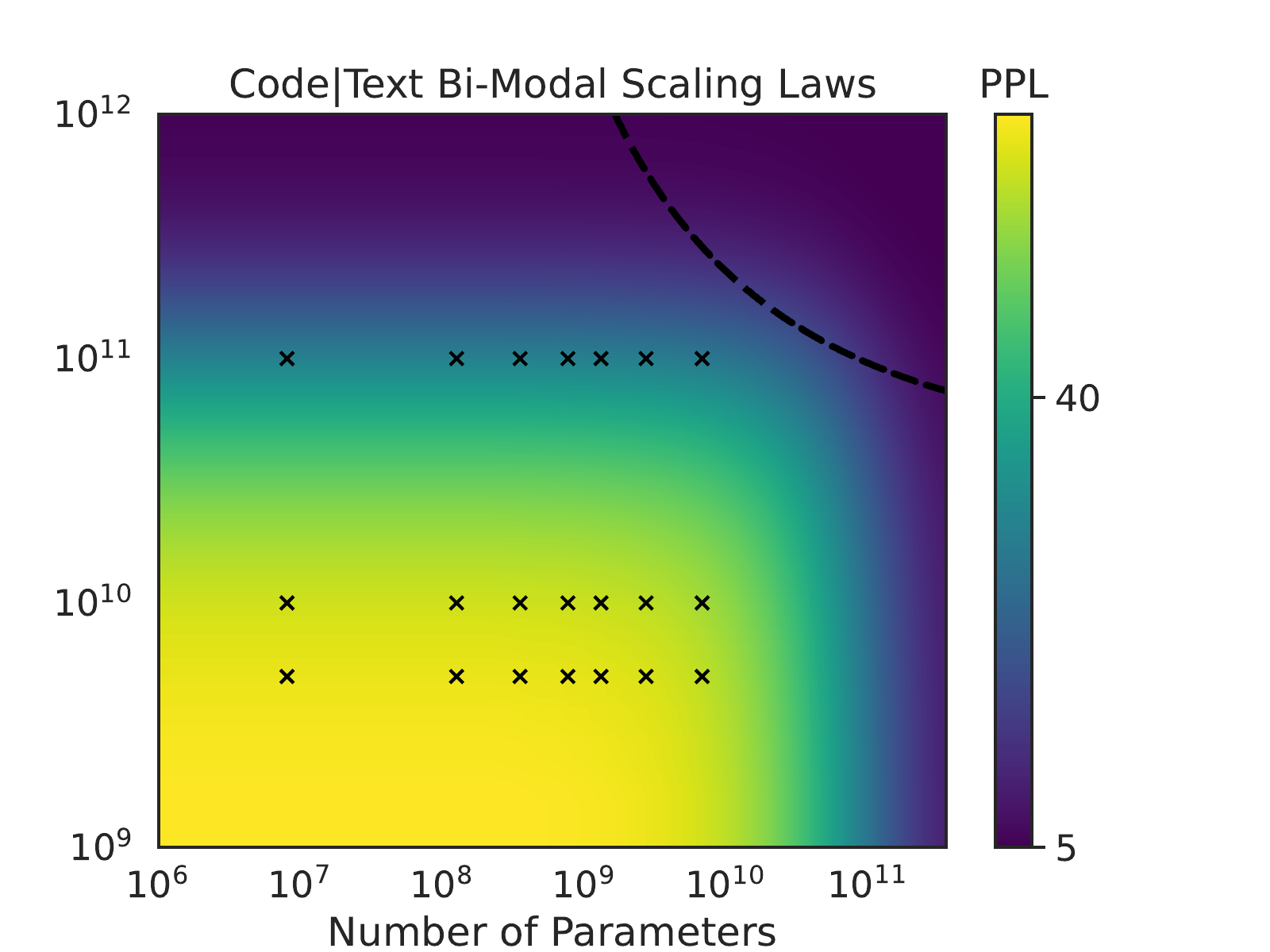}
        \end{minipage}
        \label{fig:comp_barrier}\caption{Bi-modal scaling law extrapolation and the predicted competition barrier for a subset of our bi-modal experiments.}
    \end{figure}

    %We note that for, the models trained for our empirical scaling law fitting, we do not see any modality synergy, but we see competition. 
    Given these laws, we can now make predictions about what scale will be required to overcome modal competition and achieve synergy from training on each pair of modalities. 
    By modality competition, we refer to the empirical phenomena of two modalities performing worse than if we trained two individual models on the same number of per-modality tokens. By synergy, we mean the inverse. 
    We can define the notion of synergy formally through our scaling laws. If
    \begin{equation}
        \mathcal{C}_{i,j} > \frac{A_{i,j}}{N^{\alpha_{i,j}}} + \frac{B_{i,j}}{|D_{i}| + |D_{j}|^{\beta_{i,j}}}
    \end{equation}
    we are reducing the loss beyond the independent modeling of the modalities and therefore are synergistic; otherwise, we say the modalities are in competition. When both sides of the inequality are equal, we call this the competition barrier for the two modalities. We present our extrapolated scaling laws with the predicted competition barrier in Figure~\ref{fig:comp_barrier}.
    
    We can then find the compute-optimal model size and token count that breaks the competition barrier by minimizing a compute cost over the competition barrier. We select the approximation from \citet{kaplan2020scaling}.
    \begin{align}
        \begin{split}
            \min_{N, |D|} &6ND\\& \textrm{ s.t. } \mathcal{C}_{i,j} = \frac{A_{i,j}}{N^{\alpha_{i,j}}} + \frac{B_{i,j}}{|D_{i}| + |D_{j}|^{\beta_{i,j}}}
        \end{split}
    \end{align}

    For the \texttt{Speech|Text} coupling, the predicted compute optimal parameters are $N=28.35\textrm{B}$ and $D=45.12\textrm{B}$%, we provide the compute-optimal predictions for all other couplings in \S\ref{sec:app_co_pred}. 
    
\begin{figure}
        \centering
        \includegraphics[width=0.75\linewidth]{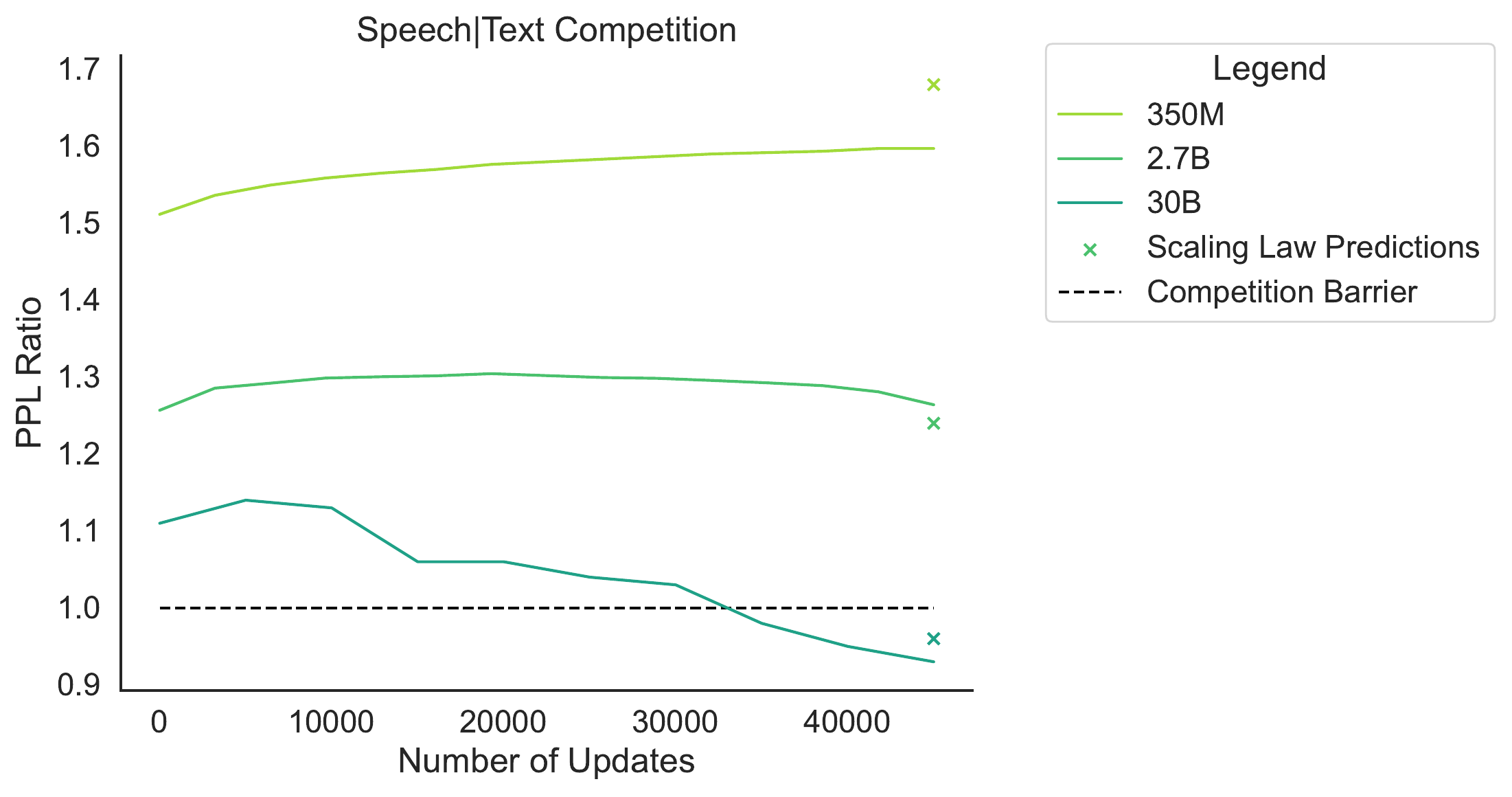}
        \caption{We plot $\frac{0.5 * \left(\mathcal{L}(N, \texttt{Text}) + \mathcal{L}(N, \texttt{Speech})\right)}{\mathcal{L}(N, \left[\texttt{Speech}, \texttt{Text}\right])} $ throughout the training process. If this ratio is below 1, we have broken through the competition barrier. Additionally, we add the predictions for the final ratio as predicted from our scaling laws.}
        \label{fig:breaking_comp_barrier}
    \end{figure}

    To test this hypothesis, we select the closest architecture available from \citet{OPT}, which is the 30B parameterization and 50B tokens, slightly above the predicted data regime, to cover any error in our approximation. We train three models a, 350M, 2.7B, and 30B models on either \texttt{Speech}, \texttt{Text}, or \texttt{Speech|Text}. We plot the ratio of the average of the \texttt{Speech} and \texttt{Text} models perplexity per timestep by \texttt{Speech|Text} perplexity, the competition barrier and predictions from our scaling laws in Figure~\ref{fig:breaking_comp_barrier}. As we see, the prediction does hold, and we achieve a model that crosses the competition barrier. Further scaling is likely to further improve the synergy, but we leave this exploration to future work.
    
\section{Emergent Phenomena}

We observed a number of emergent behaviors during training, many of which can be predicted from the modality-specific constants in our scaling laws. We briefly document these behaviors here; each is potentially worthy of study in future work. 

\begin{figure}[h]
    \includegraphics[width=0.45\textwidth]{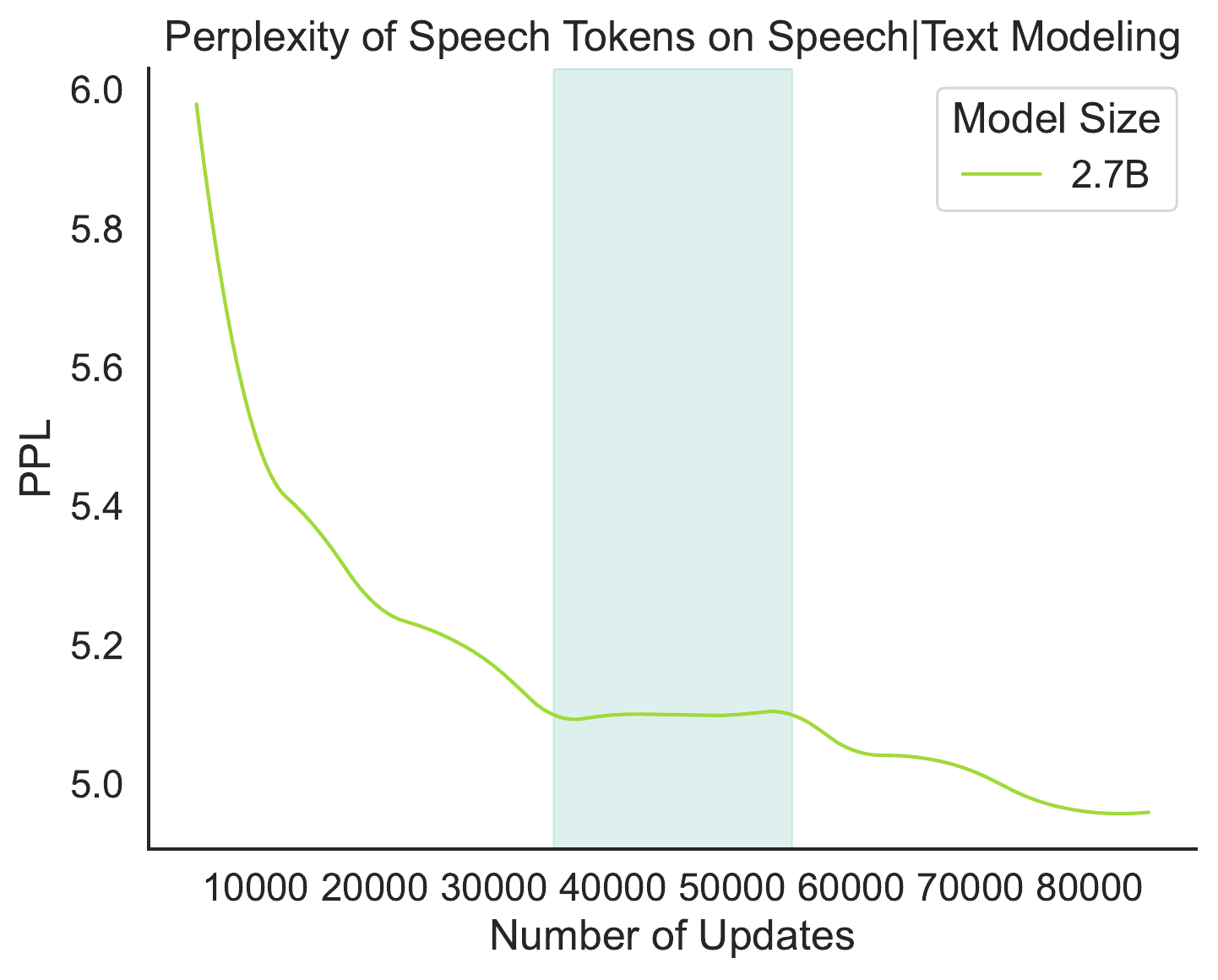}
    \includegraphics[width=0.45\textwidth]{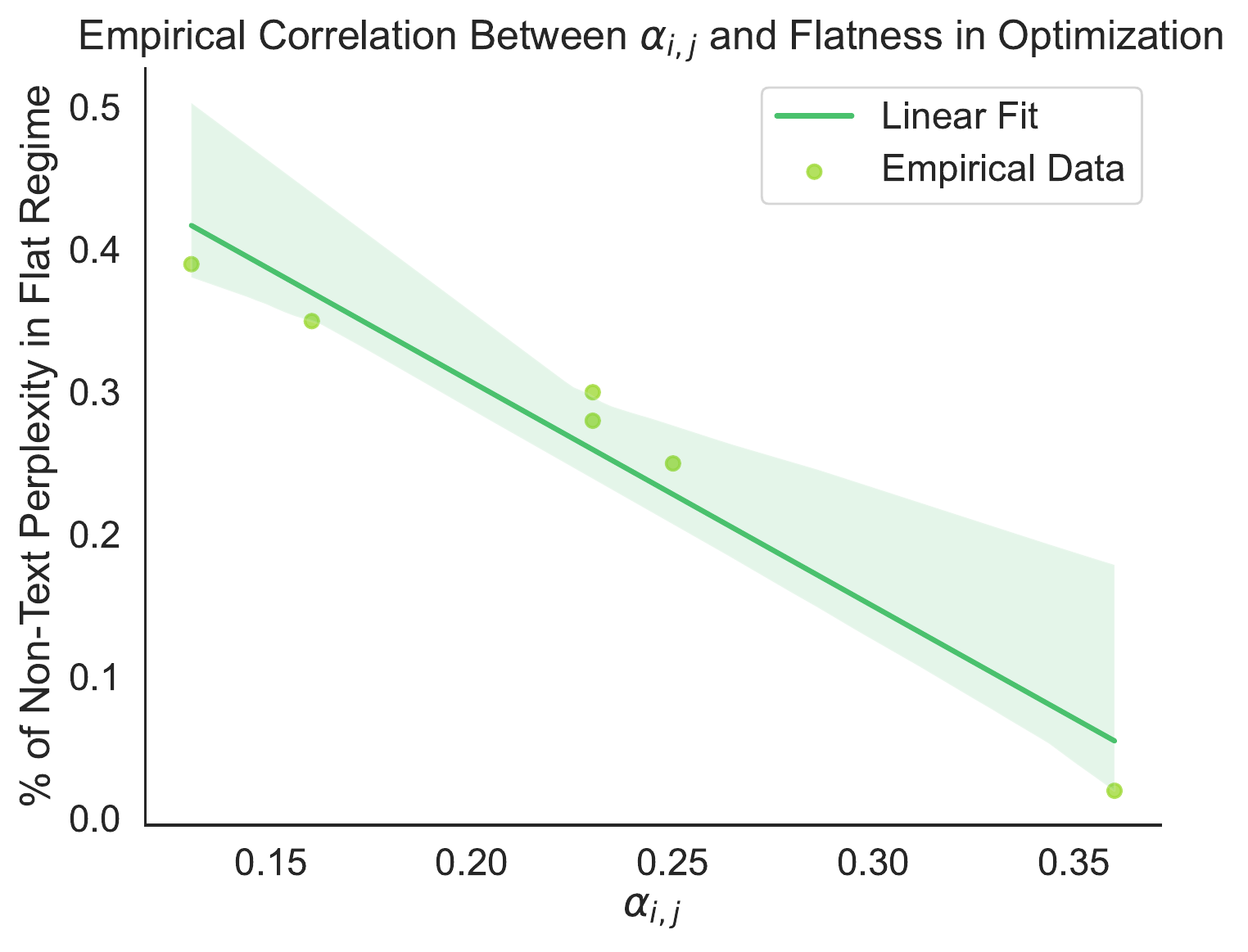}
    \caption{\textbf{Left:} Example run showing the perplexity on only the speech tokens of a 2.7B run over the \texttt{Speech|Text} dataset. We highlight a region where roughly for 15000 steps perplexity for speech flattened. \textbf{Right:} Correlation between the mixed-modal $\alpha_{i,j}$ parameter and the percent of non-text perplexity that are within a flat regime in the 6.7B model regime.}\label{fig:int_coord_ascent_sample}
\end{figure}

\begin{figure}[h]
    \centering
    \includegraphics[width=0.45\textwidth]{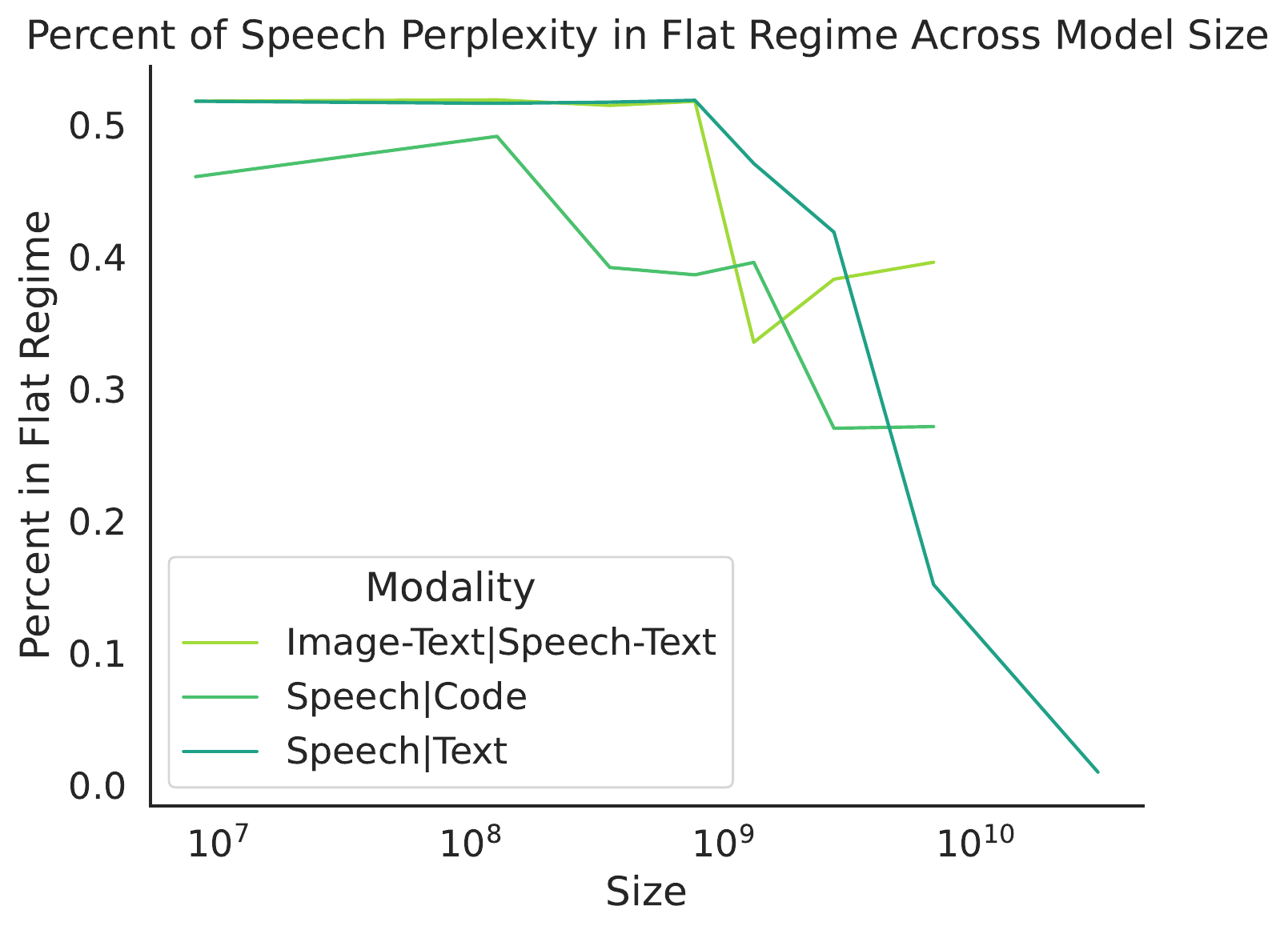}
    \includegraphics[width=0.45\textwidth]{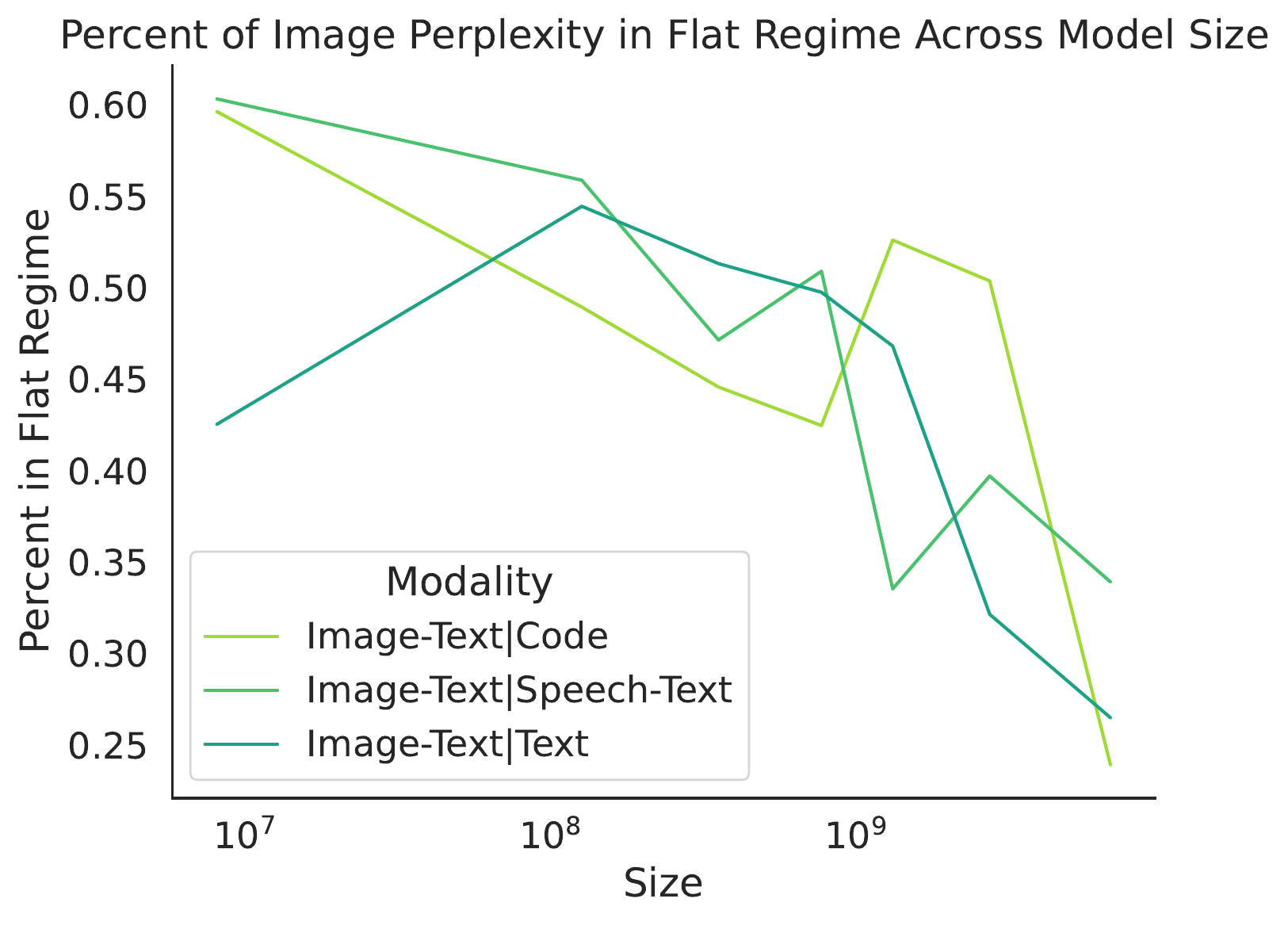}
    \caption{Percent of the submodality that exhibits flatness, where flatness is defined as an area of the training curves where loss does not decrease. We present these plots for speech and image perplexity within the bi-modal couplings that contain them.}
    \label{fig:int_coord_ascent}
\end{figure}

\begin{phenomena}
    \label{phenomena:coord_ascent}
    \textbf{Intermittent Coordinate Ascent Like Training:} Different source modalities in a multi-modal setting are optimized at different paces, with some modalities even pausing their training progression for a significant amount of steps.
\end{phenomena}
    When looking at average perplexity over the dataset, the training dynamics are always consistently smooth and somewhat monotonically decreasing (Figure~\ref{fig:monomodal_scaling}). But looking at the sub-perplexities of the modalities shows a different picture; certain modalities flatten out during training (see left figure in Figure~\ref{fig:int_coord_ascent_sample}). 
In Figure~\ref{fig:int_coord_ascent}, we plot the percent of the submodality that exhibits flatness, where flatness is defined as an area of the training curves where loss does not decrease (we do not count the warm-up period of optimization as part of this percentage).

\begin{phenomena}
    \label{phenomena:coord_ascent_scale}
    \textbf{Rate of Phenomena~\ref{phenomena:coord_ascent} Diminishes Past A Certain Scale: } The rate of intermittent coordinate ascent-like training is correlated with scale ($N$) and $\alpha_{i,j}$.
\end{phenomena}
Most of this intermittent coordinate ascent-like training can be reduced by simply increasing the model size. Intuitively, this makes sense as the increased functional approximation space should give the models enough capacity to simultanouesly optimize all of the modalities (Figure~\ref{fig:int_coord_ascent}). 
Additionally, we discover that the empirically found $\alpha_{i,j}$, which describes the functional approximation cost across two modalities, is highly correlated with the uni-modal optimization flatness in the training regime.
We found no correlation between $\beta_{i,j}$ and optimization flatness.

\begin{figure}[h]
    \centering
    \includegraphics[width=0.45\textwidth]{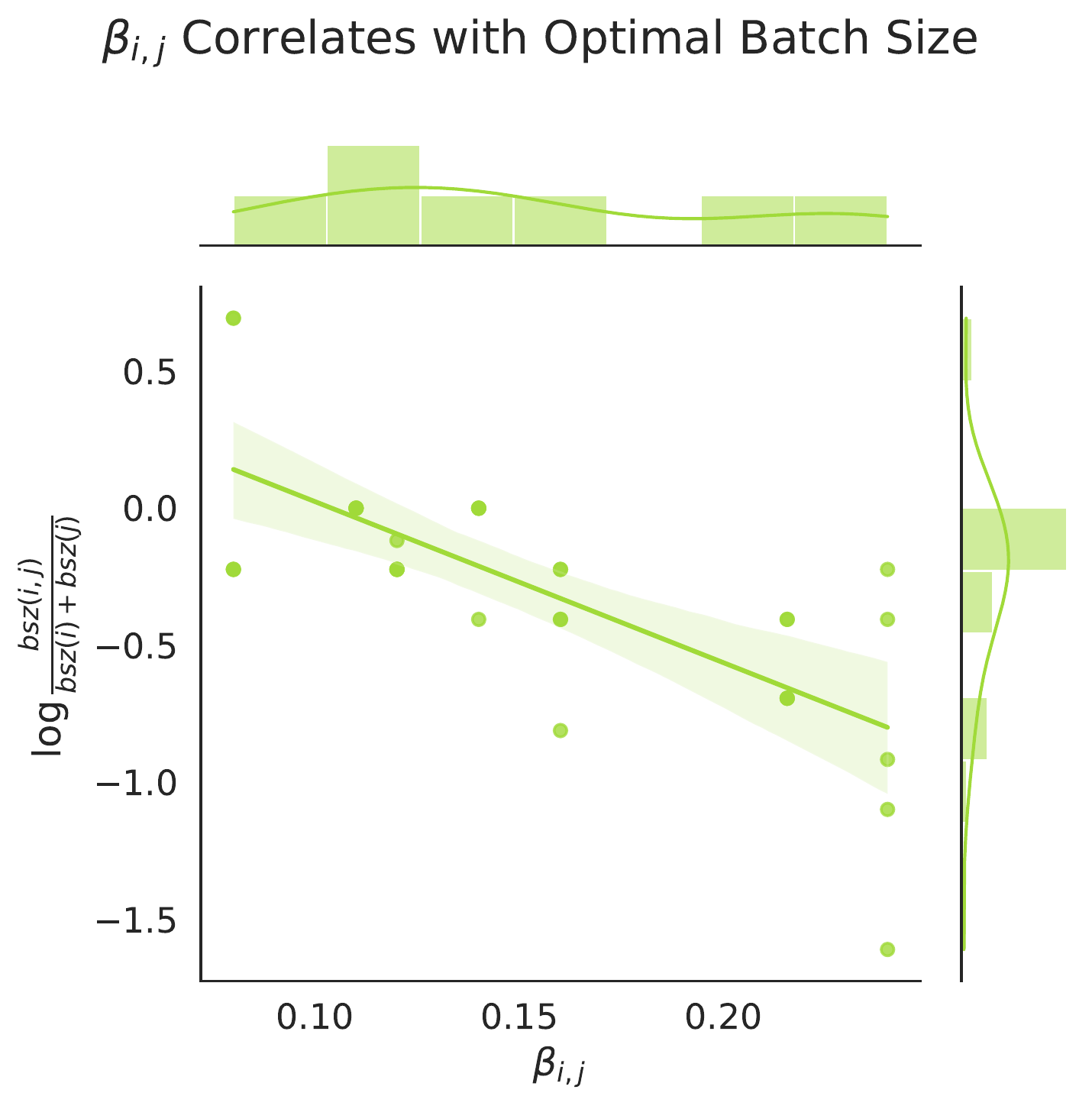}

    \includegraphics[width=0.45\textwidth,valign=t]{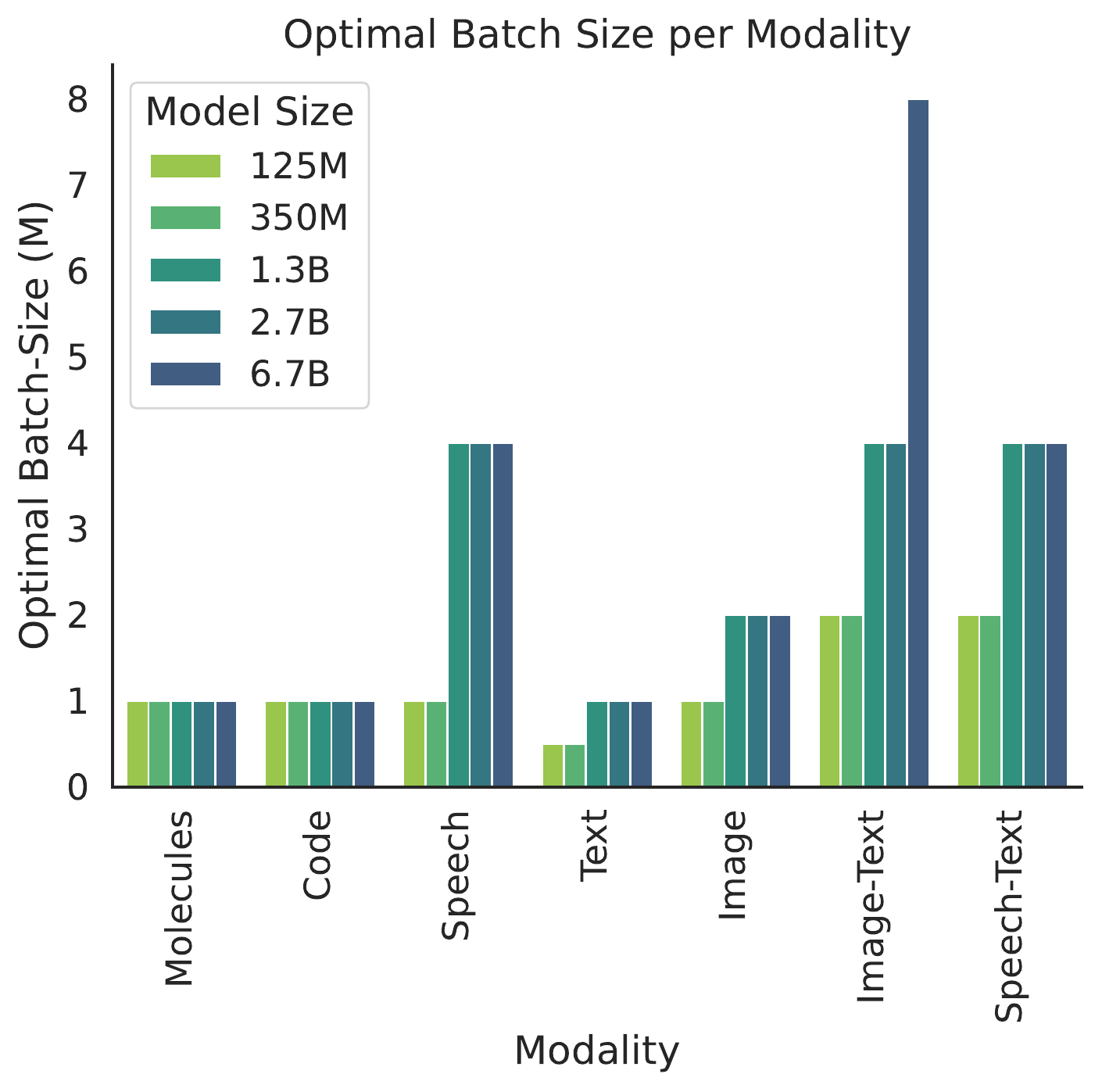}
    \includegraphics[width=0.43\textwidth,valign=t]{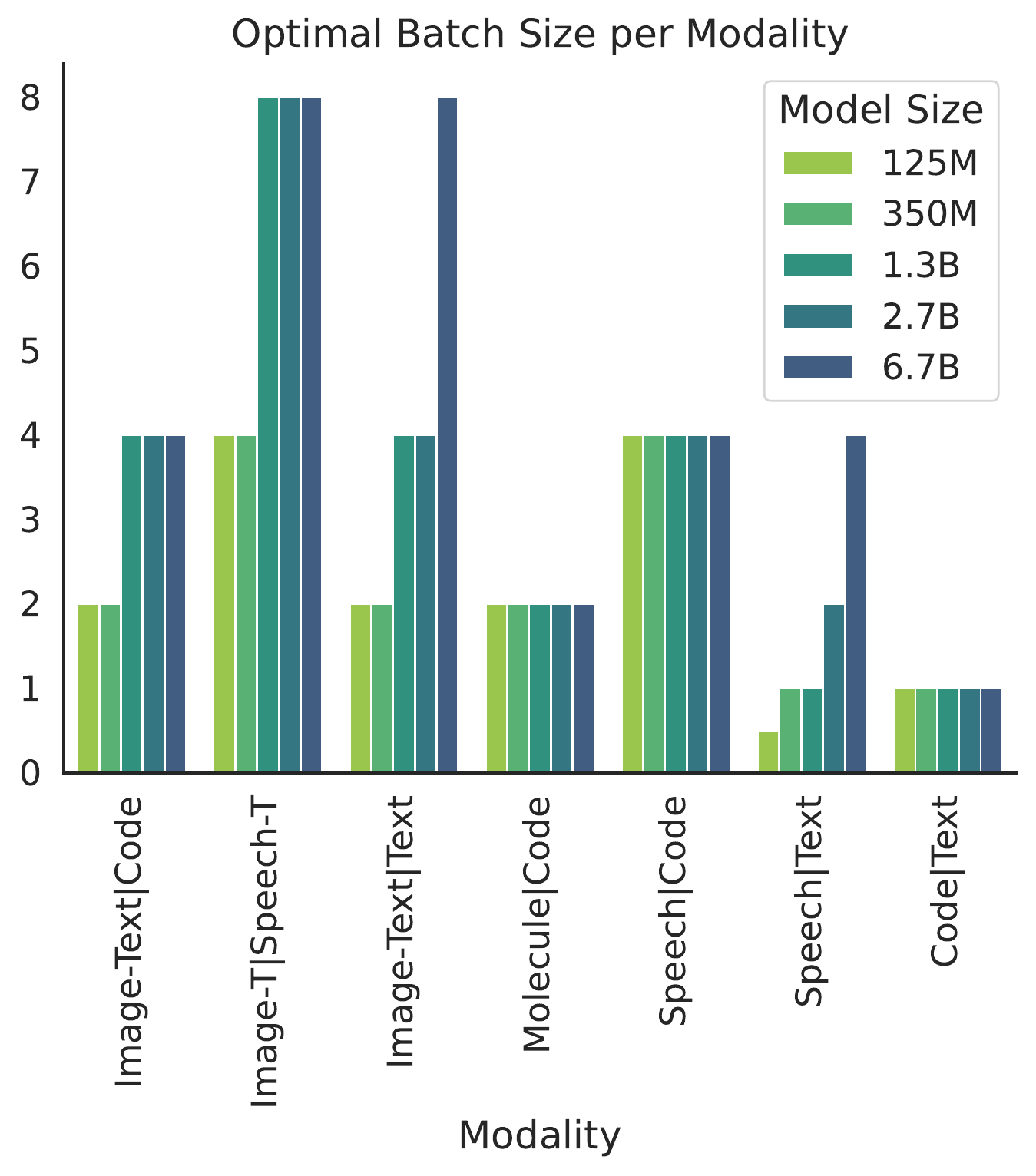}
  \label{sec:batch_size_exploration}\caption{\textbf{Bottom:} Optimal batch-size per modality and modality couplings across model sizes. \textbf{Top:} The logarithm of the ratio between the optimal batch size for the entire dataset and the sum of the optimal batch sizes for the individual sub-datasets plotted against the $\beta_{i,j}$ values of the scaling laws that were identified in the previous section.}
\end{figure}

\begin{phenomena}
    \label{phenomena:batch_size} 
    \textbf{Optimal Batch Size for Modalities $i$ and $j$ is Correlated with $\beta_{i,j}$}    
\end{phenomena}
We fixed the batch size to 1M tokens, but the question of the optimal batch for each modality and modality coupling remains. We train four versions of all over a subset of models, overall modalities, and selected couplings of modalities with batch sizes 1M, 2M, 4M, and 8M over 5B tokens, with the exception of modalities that contain \texttt{Text} for which we add 0.5M batch size experiments. We use the same training regime as mentioned in \S~\ref{sec:training_proc}.
We present our results in Figure~\ref{sec:batch_size_exploration}. Additionally for the bi-modal coupling experiments we plot $\log$ of the ratio between the optimal batch size and the sum of the optimal batch sizes for the sub-datasets against the $\beta_{i,j}$ of the discovered scaling laws in \S~\ref{sec:bimodal_scaling}. We found no correlation between $\alpha_{i,j}$ and optimal batch-size.

\begin{figure}[h]
    \centering
    \includegraphics[width=0.45\textwidth]{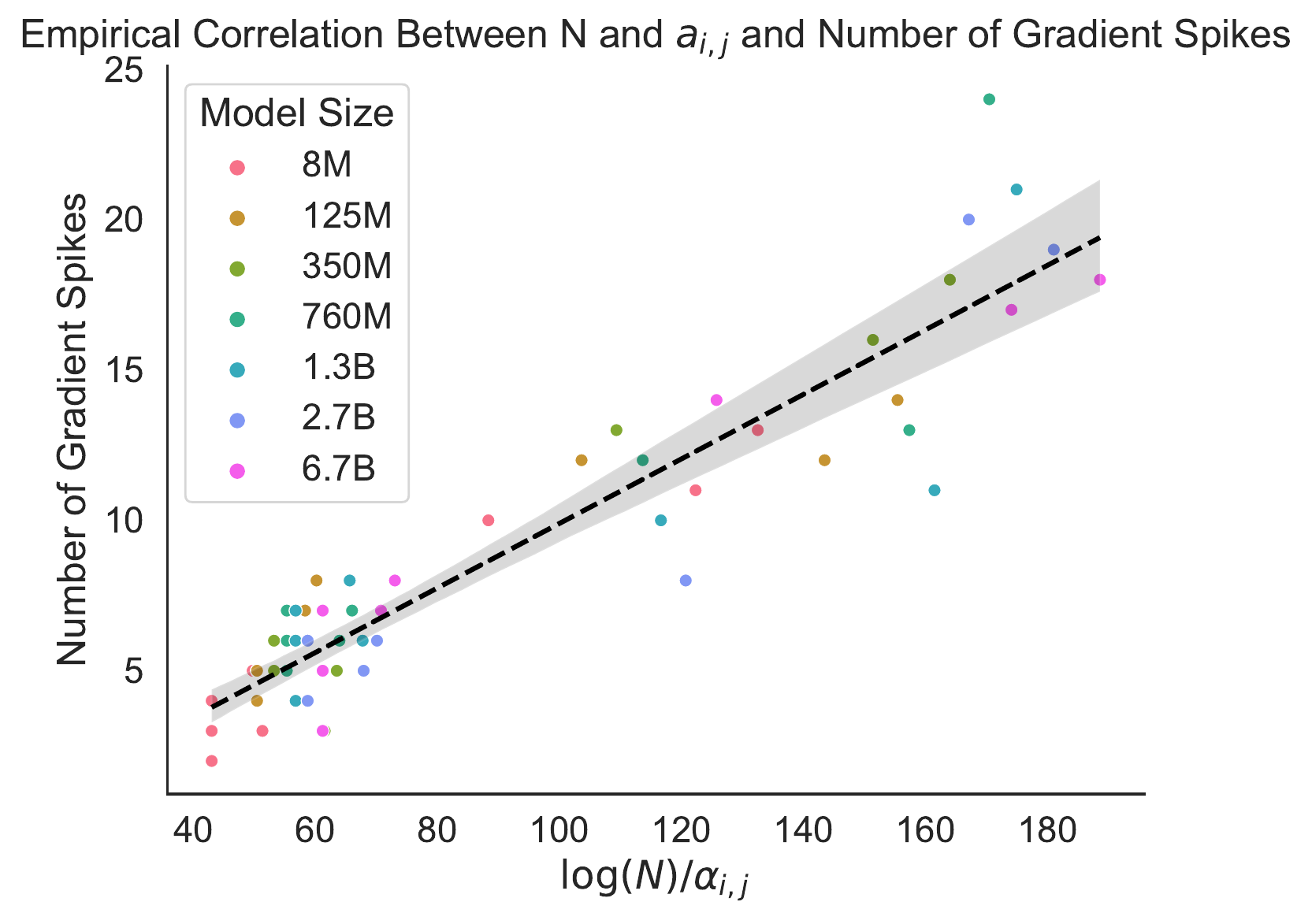}
  \label{sec:gradient_spikes}\caption{We plot $\log(N) / \alpha_{i,j}$ against the number of gradient spikes that occurred for the respective experiments.}
\end{figure}

\begin{phenomena}
    \label{phenomena:deterioting_training_dynamics} 
    \textbf{Rate of Deteriorating Training Dynamics is Correlated with $\alpha_{i,j}$ and $N$}    
\end{phenomena}
The stability of training can be captured by looking at the total count of gradient norm spikes throughout the lifetime of the training. A large number of gradient spikes can indicate a poor training setting, from selecting the wrong learning rate or batch size to having low-quality data. Additionally, larger models tend to be harder to stabilize, reflecting in a larger amount of gradient spikes. We hypothesize that lower values of $\alpha_{i,j}$, reflecting higher competition between modalities, will correlate with more gradient norm spikes. We present the empirical correlation between $\log(N) / \alpha_{i,j}$ and \# of Gradient Norm Spikes in Figure~\ref{sec:gradient_spikes}. We see a highly predictive relationship between model size ($N$) and the rate of mixed-modal competition ($a_{i,j}$ to the stability of the training run. We found no correlation between $\beta{i,j}$ and the \# of gradient norm spikes.

\section{Conclusion}
    We have provided extensive experimentation and analysis into the scaling properties of mixed-modal generative models. By developing a scaling law that reflects the contributions of individual modalities and the interaction between them, we have gained a deeper understanding of scaling mixed-modal models and the training dynamics of these models. Our findings also include a set of empirical phenomena observed during the training process and training dynamics that can be primarily explained through various interaction terms in our newly proposed scaling law. Additionally, we have developed guidelines for selecting critical hyper-parameters based on our scaling law, providing a valuable tool for practitioners in the field. Overall, our research has advanced the knowledge and understanding of mixed-modal generative models and will help develop unified models that can handle multiple modalities simultaneously.
\section{Acknowledgements}
We thank Hrant Khachatrian and Hrayr Harutyunyan for their discussions about the exact formulation of the mixed-modal scaling laws. We also thank Adam Polyak and Oran Gafni for training the Make-A-Scene tokenizer used in this work, and Rich James for editing the paper.
 
\bibliography{iclr2023_conference}
\bibliographystyle{iclr2023_conference}

\appendix
\section{Appendix}
\subsection{Model Architecture}

All models are trained with pre-norm and using ReLU activation. We apply a dropout of 0.1 throughout, but we do not apply any dropout to embeddings. We also use weight decay of 0.1. To initialize the weights, we use a variant based on Megatron-LM codebase, which involves using a normal distribution with a mean of zero and a standard deviation of 0.006. We truncate this normal distribution within two standard deviations and observed substantial gain in both training stability and performance.

\begin{table}[h]
\centering
\begin{tabular}{lrrrrrrr}
\toprule
Model & \#L & dmodel & \#H & dhead & Batch Size & LR       & Context Length \\ \midrule
8M    & 4   & 128    & 2   & 64    & 1M  & 1.00E-03 & 2048     \\
125M  & 12  & 768    & 12  & 64    & 1M  & 6.00E-04 & 2048     \\
350M  & 24  & 1024   & 16  & 64    & 1M  & 3.00E-04 & 2048     \\
760M  & 24  & 1536   & 16  & 96    & 1M  & 2.50E-04 & 2048     \\
1.3B  & 24  & 2048   & 32  & 64    & 1M  & 2.00E-04 & 2048     \\
2.7B  & 32  & 2560   & 32  & 80    & 1M  & 1.60E-04 & 2048     \\
6.7B  & 32  & 4096   & 32  & 128   & 1M  & 1.20E-04 & 2048     \\
30B   & 48  & 7168   & 56  & 128   & 1M  & 1.00E-04 & 2048    \\
\bottomrule
\end{tabular}
\label{table:opt_model_desc}
\caption{Model architecture details. We report the number of layers (\#L), the embedding size ($\text{d}_\text{model}$), the number of attention heads (\#H), the dimension of each attention head ($\text{d}_\text{head}$), batch size, learning rate (LR) and context length (\# of tokens).}
\end{table}
\subsection{Causal Masked vs. Causal Objective}
\label{sec:cm_vs_c}
    We measure the impact of the choice of objective by conducting an additional scaling law on our \texttt{Speech} and \texttt{Text} datasets on the standard (causal) language modeling objective. Everything is kept constant except for the objective, including the training procedures. We present the empirically fit scaling law parameters in Table~\ref{tab:causal_vs_causal_masked}.
    \begin{table}[h]
        \centering\small
        \begin{tabular}{lrrrrr}
        \toprule
        {} & A & B & E & $\alpha$ & $\beta$ \\
        \midrule
        Speech (CM3) & 154.45 & 205.10 & 3.02 & 0.31 & 0.24 \\
        Speech (Causal) & 164.12 & 201.00 & 3.01 & 0.30 & 0.24 \\ \midrule
        Text (CM3) & 492.51 & 1987.40 & 2.42 & 0.18 & 0.22 \\ 
        Text (Causal) & 485.16 & 1859.32 & 2.45 & 0.17 & 0.23 \\
        \bottomrule
        \end{tabular}
        \caption{Uni-Modal scaling law parameters fit to Equation~\ref{eq:chinchilla_scaling_law} 
 for both causal (standard language modeling) and causal masked (CM3 objective from \citet{CM3}).}
        \label{tab:causal_vs_causal_masked}
    \end{table}

    Note that both objectives optimize the joint probability of tokens; therefore, if there was a significant difference in our perplexity, we should expect to see it reflected in a difference in scaling law parameters. Instead, we see that the scaling laws seem to be close to identical, with whatever minor differences within the error of our approximation.
    
\subsection{Tokenization}
\newcommand{\vqgan}{\text{VQGAN}}
\newcommand{\multigen}{\text{VQGAN}_\text{MAS}}
\newcommand{\multigenS}{\text{VQGAN}_\text{MAS} 256}
\newcommand{\multigenL}{\text{VQGAN}_\text{MAS} 512}

\subsubsection{Quality of Image Tokenization}
\label{sec:image_tokenization}

Modeling long-range dependencies with raw pixel input of an image (for example, total sequence length for a 256-pixel image in RGB form is 196608) is non-trivial, especially with transformers, which in their vanilla form scale poorly with sequence length. Recently, Vector Quantized Variational autoencoders (VQ-VAE, or Discrete-VAE) have been proposed, which learn discrete image representations, allowing a later generative model to generate images in the discrete latent space, just like a standard language model. VQ-VAE reduces the context size of a transformer by a factor of $3*X^2$ ($X$ is the spatial reduction rate, and 3 is the number of image channels), where information loss is unavoidable. VQ-VAE is trained to optimize the evidence lower bound of distribution of data. \cite{taming} introduced $\vqgan$, which improves upon VQVAE by introducing an adversarial loss produced by a discriminator, reconstructing images with much higher quality. Recently, \cite{make_a_scene} trained a new image tokenizer with a better training objective focusing on faces or objects, which is adopted for this work and denoted as $\multigen$. To be most effective in the later language model stage, the image tokenizer must represent an image effectively. The correlated decoder must reconstruct the generated image tokens into high-quality image data. 
We benchmark the following pre-trained tokenizers on these properties:
\begin{itemize}
\item $\vqgan (fx, y)$ with different spatial reduction rate $fx$ and diffent vocab size $y$. For example, for a 256px image, 256 tokens will be created with a $\vqgan (f16)$ tokenizer and 1024 tokens with a $\vqgan (f8)$ tokenizer.
\item Our $\multigenS$ and $\multigenL$ use $f8$ and $f16$ spatial reduction, respectively, and have an 8192 vocab size. Our $\multigenS$ is trained with a face-aware loss with the help of a pre-trained face embedding model. Our $\multigenL$ is trained with face+object aware loss with extra downsampling and upsampling layer in the encoder and decoder to reconstruct images with higher resolution. Note, Our $\multigenL$ with 512x512 image input compresses the image to 1024 tokens, thanks to the downsampling layer.
\end{itemize}

One way to quantify the realism captured by these models is to compute Fréchet Inception Distance (FID) scores of reconstructed images w.r.t. the inputs (R-FIDs). Table \ref{tab:reconstruct} shows R-FIDs when reconstructing the whole validation split of the ImageNet dataset. For an image with a 256-pixel resolution, reducing spatial reduction rate or increasing visual vocab size can help achieve lower R-FIDs. Our $\multigenS$ model is superior to its counterpart with the same spatial reduction rate and vocab size, demonstrating the effectiveness of the extra face-aware loss. Interestingly, $\multigenS$ gets a higher R-FID than $\multigenL$, consistent with the result in the original paper. We are also interested in understanding how much the reconstruction process can retain information and if we lose critical image information. We benchmark the representation power via classification accuracy with a pretrained model. We first reconstruct all the images in the ImageNet validation set with different tokenizers, similar to R-FID computation. Then a trained pretrained classifier on ImageNet is used to run inference on the original and reconstructed images. The classification accuracy of original images with 256 or 512-pixel resolution is 81.56 and 82.89, respectively. The accuracy@1 on the reconstructed images and their gap with the raw images are reported in Table ~\ref{tab:reconstruct}. Images reconstructed by the $\multigenS$ can best maintain the original information with less than 2 percent degradation in accuracy. 

\begin{table}[h]
\centering
\begin{tabular}{lrrccc}
\toprule
\multicolumn{1}{l}{Tokenizer} & \multicolumn{1}{l}{Spatial Reduction} & Vocab & Token Counts & R-FID         & Accuracy@1             \\
\midrule
\multicolumn{6}{l}{256x256 px}                                                                                                                        \\
   \midrule       
$\vqgan$                         & 16                                    & 1024                               & 256          & 7.94          & 69.25 (-12.4)          \\
$\vqgan$                         & 16                                    & 16384                              & 256          & 4.98          & 73.2 (-8.45)           \\
$\vqgan$                         & 8                                     & 8192                               & 1024         & 1.49          & 79.64 (-2.01)          \\
$\vqgan$                         & 8                                     & 16384                              & 1024         & 1.14          & 79.47 (-2.18)          \\
$\multigenS$                   & 8                                     & 8192                               & 1024         & \textbf{0.87} & \textbf{79.83 (-1.82)} \\
\midrule
\multicolumn{6}{l}{512x512 px}          \\
\midrule                                 
$\multigenL$                   & 8                                     & 8192                               & 1024         & 1.43          & 79.87 (-3.02)         \\
\bottomrule
\end{tabular}
\label{tab:reconstruct}
\caption{R-FID and accuracy with images reconstructed by a selection of image tokenizers}
\end{table}

\begin{figure}[th]\centering
\includegraphics[width=1\linewidth]{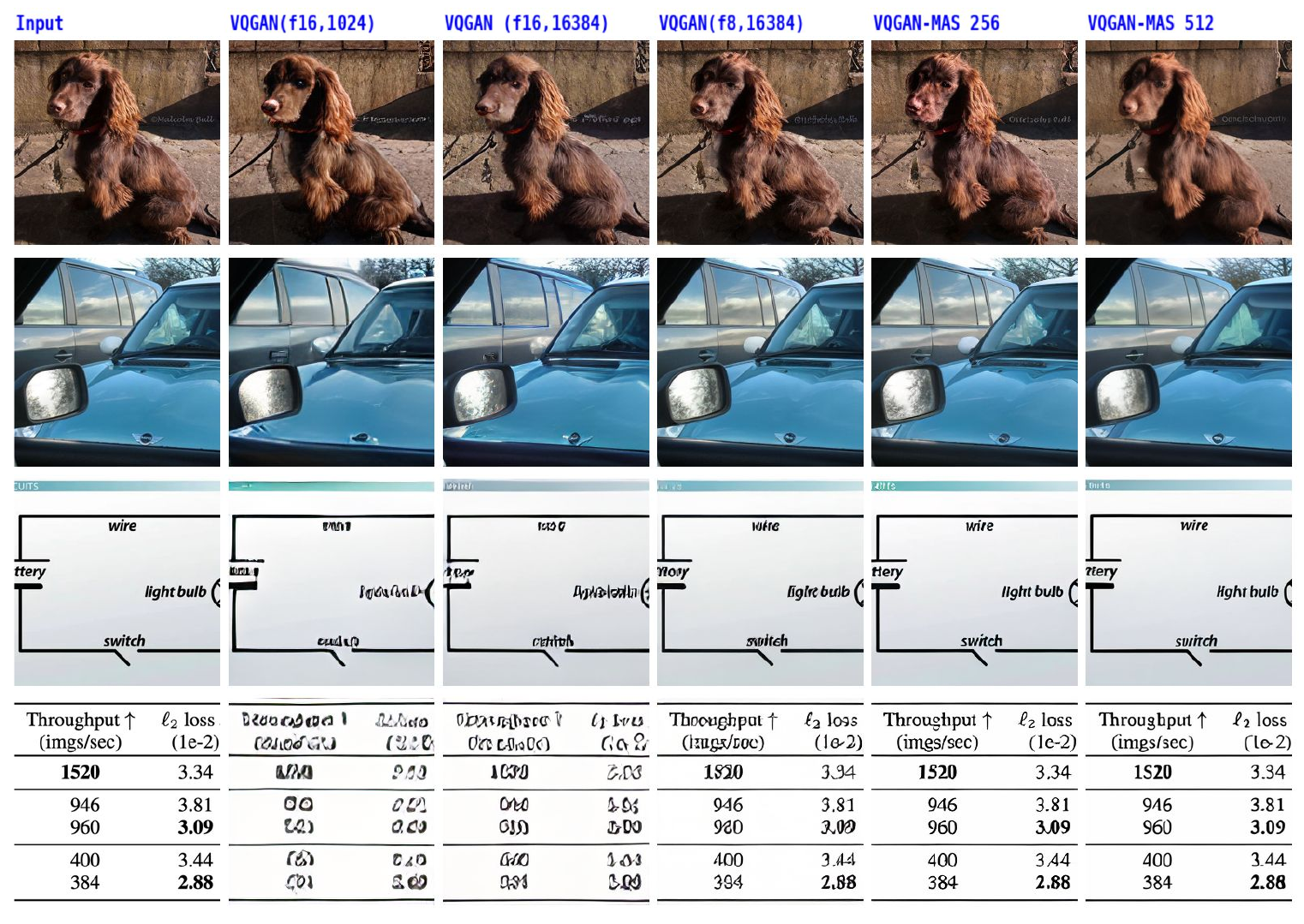}
\caption{Reconstructed images across a selection of different image tokenizers.}
\label{fig:reconstructed}
\vspace{1pt}
\end{figure}

For qualitative comparison, we give all tokenizers except $\multigenL$ an image with 256x256 pixels. $\multigenL$ reconstructs a 512-pixel resolution image and resizes it to 256-pixel resolution for plotting purposes. $\vqgan~(f16)$ a s produce 256 tokens, while $\vqgan~(f8)$ and $\multigen$ models produce 1024 tokens. 

We randomly sample two images from ImageNet (top 2 rows in Figure \ref{fig:reconstructed}). All reconstructed images can maintain vital information about the image and the textures. With a high reduction rate (192), $\vqgan$ with $f16$ spatial reduction can not reproduce every detail of its input but tends to hallucinate parts of it, for example, the eye and the tail of the dog in row 1 and the mirror of the blue car in row 2. By increasing the vocab size, more realistic images can be generated. With a decreased compression rate, the $\vqgan~(f8)$ model and $\multigen$ produce much more realistic reconstructed images. For example, in row 2, the door handle, the clouds, and the mirror's tree are successfully reconstructed with great detail. 

Lastly, we reconstruct images from a textbook (row 4 in Figure \ref{fig:reconstructed}) or screenshots of tables from scientific papers (row 5 in Figure \ref{fig:reconstructed}). All models struggle to reconstruct the original image, except $\multigen$ models. Figure \ref{fig:reconstructed} shows impressive results by $\multigen$ models that all text and numbers are human readable. $\multigenS$ produces sharper edges while $\multigenL$ smooths things out. 

\begin{figure}[th]\centering
\includegraphics[width=1\linewidth]{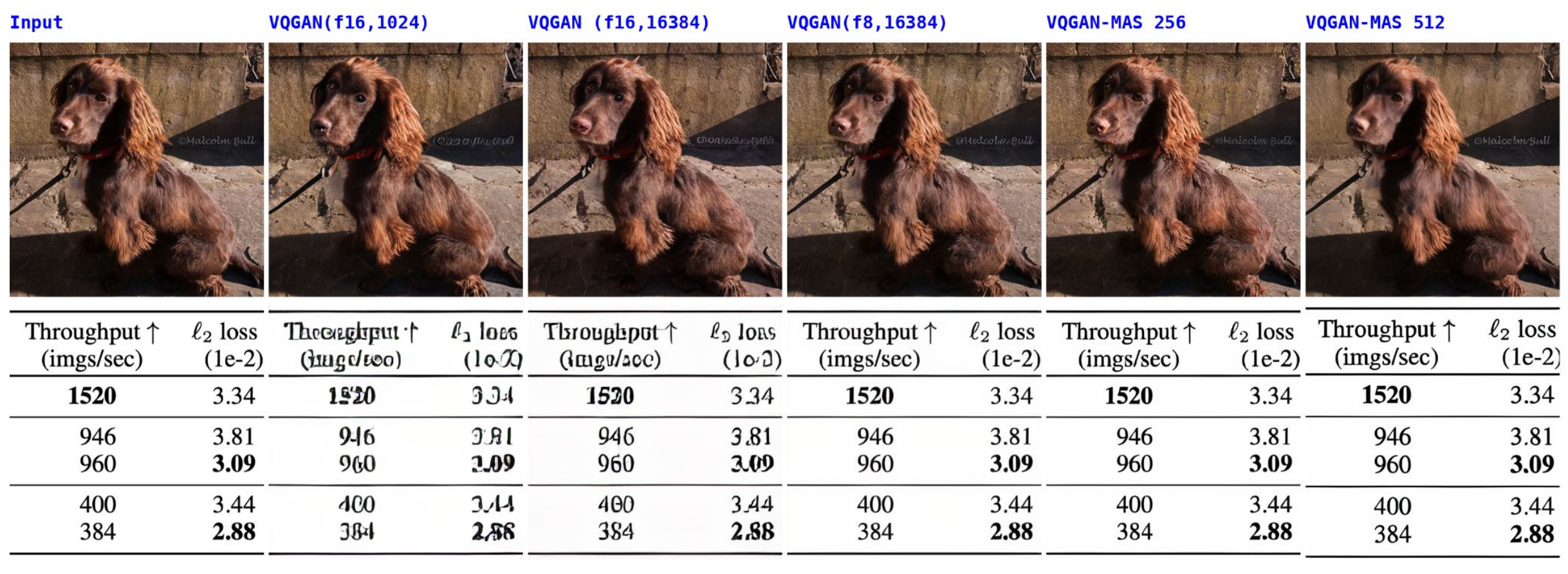}
\caption{Reconstructed images across a selection of different image tokenizers (image size 512)}
\label{fig:reconstructed512}
\vspace{1pt}
\end{figure}

From all the above examples, the reduced spatial reduction is effective for better tokenization; however, it results in a longer token sequence. Another way to increase image representation is to increase the pixel numbers of images. We reconstruct images with a size of 512x512 for $\multigenL$ in Figure \ref{fig:reconstructed512}. $\vqgan (f16)$ produce 1024 tokens, while $\vqgan (f8)$ and $\multigen$ models produce 4096 tokens. $\multigenS$ in Figure \ref{fig:reconstructed} outperform $\vqgan$ (f16) by a big margin. With the same token budget, decreasing spatial reduction is more effective than increasing image pixels. 

\subsubsection{Details of Speech Tokenization}
\label{sec:speech_tokenization}
    We use the \textsc{Base} HuBERT model in our work. 
    This model comprises a convolutional encoder and 12 layer Transformer, each with an embedding dimension of 768, a feed-forward layer dimension of 3072, and 12 self-attention heads. 
    Pre-training of the model has been performed on 32 GPUs over three iterations, with 400K updates per iteration. 
    The training data consists of 221K hours of unlabeled speech from multilingual Librispeech (MLS) ~\cite{pratap2020mls}, Common Voice (CV)~\cite{ardila2019common}, and VoxPopuli (VP)~\cite{wang2021voxpopuli} in eight languages (English, Spanish, French, German, Dutch, Italian, Polish, Portuguese). The MFCC/6-th layer feature from iteration 1 and the 9-th layer feature from iteration 2 are used as targets, with codebook sizes of 100/500/1000, respectively, following the methodology outlined in \cite{lee2021direct}. 
    
    A typical 16kHz audio with a bit depth of 16 has a bitrate of 64kbps. HuBERT encodes audio at 50Hz with a codebook size of 2000, resulting in a bitrate of 548bps. The effective compression rate is roughly 117. Our model still effectively retains speech information, as shown in Table~\ref{tab: speech-wer}. We compare the word error rate (WER) of a pretrained automatic speech recognition (ASR) model with original audio or reconstructed audio by HuBERT models. We present results with two HuBERT models, one public~\cite{hubert} version (HuBERT public) and one trained by us (HuBERT ours). WER of the original audio on LJSpeech is 2.04, the audio reconstructed by HuBERT public degrades by 0.94, while the audio reconstructed by our HuBERT only degrades it by 0.3. A similar phenomenon is observed on the LibriSpeech dataset, where our HuBERT model improves upon HuBERT public and can effectively reconstruct audio with very little information loss.

\begin{table}[h]
\centering
\begin{tabular}{lccccccc}
\toprule
Model         & PT/KM data &  Vocoder data &  \#L & K    & LJSpeech & LibriSpeech \\
\midrule
Orig audio        &            &     &      &              & 2.04     & 3.55        \\
\midrule
HuBERT public    & LS960      & LJ         & 9   & 500    & 2.98     & 12.39       \\
HuBERT ours         & MLS+VP+CV  & LJ + MLS-40h  & 12  & 2000 &  2.34     & 9.06       
\\ \bottomrule
\end{tabular}
\label{tab: speech-wer}
\caption{Word error rate (WER) on LJSpeech and LibriSpeech datasets of an petrained automatic speech recognition (ASR) model with various speech inputs (original audio, or recutructed audio by HuBERT model in \cite{hubert}, or HuBERT model in our work). We also listed the model details of two HuBERT models, including data used during pretraining (PT), k-means (KM) and vocoder, number of layers ( \#L), and number of clusters (K).}
\end{table}

\section{Credit}
\begin{itemize}
    \item \textbf{Armen Aghajanyan:} Proposed the original idea, co-authored the ablation plan, executed all the training runs and scaling law research, and was the primary writing author of the paper.
    \item \textbf{Lili Yu:} Core contributor to mixed-modal evaluations framework, drove the selection of speech/image/text tokenizer, secondary writing author of the paper.
    \item \textbf{Alexis Conneau:} Drove high-level direction, co-authored ablation plan, and collected speech datasets.
    \item \textbf{Wei-Ning Hsu:} Provided day-to-day feedback on speech-language model training, trained speech tokenizer, and tokenized all the speech data used throughout the project.
    \item \textbf{Karen Hambardzumyan:} Helped in the core design of scaling laws and writing.
    \item \textbf{Susan Zhang, Stephen Roller, Naman Goyal:} Provided support for the general training of all the models and the metaseq framework.
    \item \textbf{Omer Levy}: Provided, developed, and distilled the story for this paper. Edited paper as well.
    \item \textbf{Luke Zettlemoyer:} Provided support and feedback throughout the whole lifetime of the project. Provided help writing the paper as well as significant feedback for the paper.
\end{itemize}
\end{document}